\documentclass{article}

\usepackage{microtype}
\usepackage{graphicx}
\usepackage{subcaption}
\usepackage{booktabs} 

\usepackage{hyperref}

\usepackage[accepted]{icml2026}

\usepackage{amsmath}
\usepackage{amssymb}
\usepackage{mathtools}
\usepackage{amsthm}
\usepackage{enumitem}

\usepackage[capitalize,noabbrev]{cleveref}

\theoremstyle{plain}

\theoremstyle{definition}

\theoremstyle{remark}

\usepackage[textsize=tiny]{todonotes}

\usepackage{multirow}
\usepackage{booktabs}
\usepackage{subcaption}
\usepackage{tabularx}
\usepackage{makecell}

\icmltitlerunning{ProtoKV: Streaming Video Understanding under Delayed Query with Summary-State Memory}

\begin{document}

\twocolumn[
  \icmltitle{ProtoKV: Streaming Video Understanding under Delayed Query with Summary-State Memory}



  \icmlsetsymbol{equal}{*}

  \begin{icmlauthorlist}
    \icmlauthor{Le Tu Ngoc Minh}{equal,sch_ee}
    \icmlauthor{Jinyeong Lim}{equal,sch_ee}
    \icmlauthor{Dongsu Han}{sch_ee,sch_ai}
  \end{icmlauthorlist}

  \icmlaffiliation{sch_ee}{School of Electrical Engineering, KAIST, Daejeon, Republic of Korea}
  \icmlaffiliation{sch_ai}{Kim Jaechul Graduate School of AI, KAIST, Daejeon, Republic of Korea}

  \icmlcorrespondingauthor{Dongsu Han}{dhan.ee@kaist.ac.kr}

  \icmlkeywords{Machine Learning, ICML}

  \vskip 0.3in
]



\printAffiliationsAndNotice{\icmlEqualContribution}

\begin{abstract}
Streaming video understanding (SVU) must answer queries that arrive asynchronously while visual tokens stream continuously under strict GPU-memory and query-time latency budgets. A key challenge is delayed query: decisive cues may appear briefly, yet many subsequent updates occur before the query arrives, increasing the risk that those cues are evicted or diluted under bounded memory. We propose ProtoKV, a constant-footprint SVU memory that represents far history as a fixed-capacity summary state rather than retaining token instances. ProtoKV keeps an exact near-window KV cache and aggregates older content into a semantic–spatial prototype bank with residual statistics. At query time, each prototype is exposed through a bounded pseudo-token interface that is drop-in compatible with standard attention. Under matched budgets and comparable query-time cost, ProtoKV improves accuracy by up to 12.5 points over token-retention baselines on SVU benchmarks in the long-delay regime, with gains that grow as query delay increases.
\end{abstract}
\section{Introduction}
Streaming video understanding (SVU) systems must reason over continuous video signals in real-time \cite{rvs, ovobench, streamingbench}. Despite progress on offline benchmarks, real-world deployment requires handling continuous frames and asynchronous queries under a fixed GPU memory budget \cite{providellm}. In this setting, the model cannot assume access to the full clip nor condition computation on a known query; instead, it must maintain an evolving online state that remains immediately query-able upon demand \cite{videollm-online, rvs}.

A central difficulty in SVU is delayed query, where decisive cues (e.g., a momentary state change) appear briefly and long before a query is issued \cite{reST, ego4d}. Here, delay is not only a time gap but also the amount of post-evidence update pressure accumulated before answering. As the temporal gap between evidence and query grows, answering quality depends increasingly on what the system preserves from far history; this induces a near–far regime where recent content is kept exactly within a short window, while older history must rely on a compact representation.

Bounded-memory designs typically occupy two extremes: sliding-window baselines maintain constant memory by retaining only recent content but inevitably lose evidence once it expires \cite{streamingllm, longformer}. Conversely, offloading or retrieval systems recover fine-grained details but add significant query-time overhead and variance \cite{rekv, shadowkv}. Between these extremes lies a critical operating point for interactive systems: always-on; query-agnostic maintenance; a fixed on-device footprint usable directly at query time without reconstructing the past.

Recent work advances this operating point via online KV-cache compression under a hard memory cap for SVU \cite{infinipotv}. While this class of methods yields predictable query-time cost, online retention decisions are brittle when answers depend on rare but decisive evidence across long delays. More broadly, tying far-history to specific token instances means that once a token is dropped, its associated fine-grained cue is lost \cite{h20, snapkv}.

We propose ProtoKV, a constant-footprint SVU memory mechanism in this regime that shifts far-history representation away from token retention via a two-tier online state: (i) a short near window storing exact KV for recent frames, and (ii) a fixed-capacity far memory summarizing older history into object-centric prototypes with lightweight residual statistics. ProtoKV does not retain token-level KV for the distant past; instead, it leverages pseudo-tokens and mass-aware weighting to allow standard attention to access far-history evidence under a constant query-time footprint. This design is tailored to delayed queries depending on past event occurrence rather than instantaneous visual content. Evaluations across four streaming video benchmarks demonstrate that ProtoKV improves accuracy and degrades more gracefully than token-retention baselines as evidence becomes increasingly distant under matched memory budgets. Our code is available at \url{https://github.com/kaist-ina/ProtoKV}.

\textbf{Contributions.} (1) We introduce a far-memory mechanism for SVU that summarizes distant history into a fixed-capacity, spatiotemporally consistent prototype bank augmented with residual statistics. (2) We propose pseudo-token synthesis with mass-aware weighting, enabling standard attention to exploit far-history evidence under a bounded query-time footprint. (3) We introduce a delayed-query protocol with retrospective-query filtering to keep invariant ground truth.
\section{Background and Problem Setup}

\subsection{KV Growth as an Operational Constraint in SVU}
Streaming video understanding (SVU) requires maintaining an online state to process continuous video streams and answer arbitrary queries without replaying past frames \cite{videollm-online,providellm}. In Transformer-based VLM, this state scales linearly with time as each visual token is appended to the KV cache \cite{vit}. Given the unbounded nature of streaming, this context growth inevitably exceeds fixed deployment budgets, creating a fundamental scalability bottleneck \cite{vllm, h20}.

This pressure is amplified in video due to the high token arrival rate, quickly producing large token histories even with aggressive sampling. As a result, SVU systems must maintain an online state that supports immediate answering at query time, but whose memory cost does not scale with stream length. While simple heuristics like window truncation or token downsampling can stabilize memory \cite{streamingllm, longformer, llama_vid}, they do so by sacrificing long-range evidence or fine-grained details, respectively. Consequently, SVU demands bounded-memory mechanisms that preserve past representations in a fixed-capacity footprint while remaining compatible with standard attention at query time.

\subsection{Bounded-Memory SVU and Prior Approaches}
In contrast to offline video understanding, where the full sequence is accessible and models can compress or revisit tokens with complete context \cite{mlvu, videomme}, SVU must process frames online without knowing the stream horizon in advance. In practice, this makes it unrealistic to “buffer the whole video” before answering. We therefore focus on bounded-memory SVU under three deployment-driven constraints. First, the memory footprint must be capped by design (constant or hard-bounded). Second, the online state must be maintained query-agnostically: the system cannot tailor what it stores to a future query it has not seen yet, nor can it replay the stream at query time. Third, the stored state must be immediately consumable at query time without expensive reconstruction steps that introduce latency variance.

While some works utilize end-to-end trained compression \cite{providellm}, we focus on training-free, drop-in mechanisms that are easier to deploy across models and backbones. Prior approaches span a spectrum of trade-offs under these constraints. Sliding-window KV caching (SWA) guarantees constant memory by retaining only the most recent window, yielding predictable cost but losing evidence once it exits the window \cite{streamingllm, longformer}. Online token-retention/compression methods \cite{infinipotv} maintain a fixed KV budget by selecting, merging, quantizing, or evicting token instances over time. While they can preserve some far history, they still represent far evidence as a subset of token instances, and must continually decide which instances survive under a hard cap. Offloading/retrieval-based SVU \cite{rekv} stores history in slower tiers or retrieves it at query time to recover fine-grained details; however, the resulting query-time overhead and variance can be a practical weakness when responsiveness is central (e.g., storage/network contention or strict tail-latency targets). These trade-offs shape the operating point required for practical SVU deployments: always-on, query-agnostic maintenance of a constant-footprint on-device state that can be directly used at query time, without reconstructing the past.

\begin{figure}
    \centering
    \includegraphics[width=\linewidth]{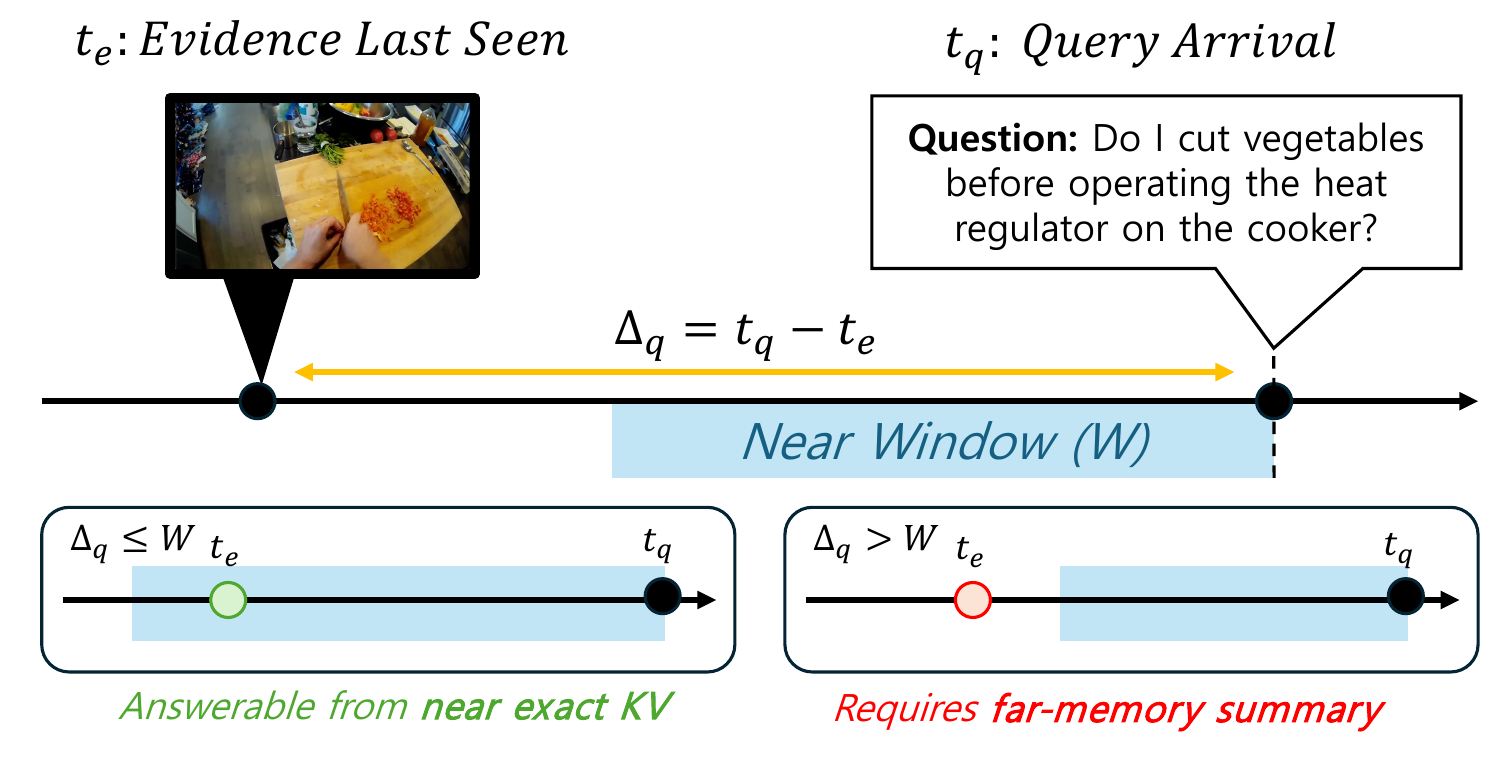}
    \caption{Query delay in streaming video understanding. When the delay exceeds the near window, answering requires information preserved in far memory.}
    \label{fig:delay}
\end{figure}

\begin{figure*}[t]
    \centering
    \includegraphics[width=0.95\textwidth]{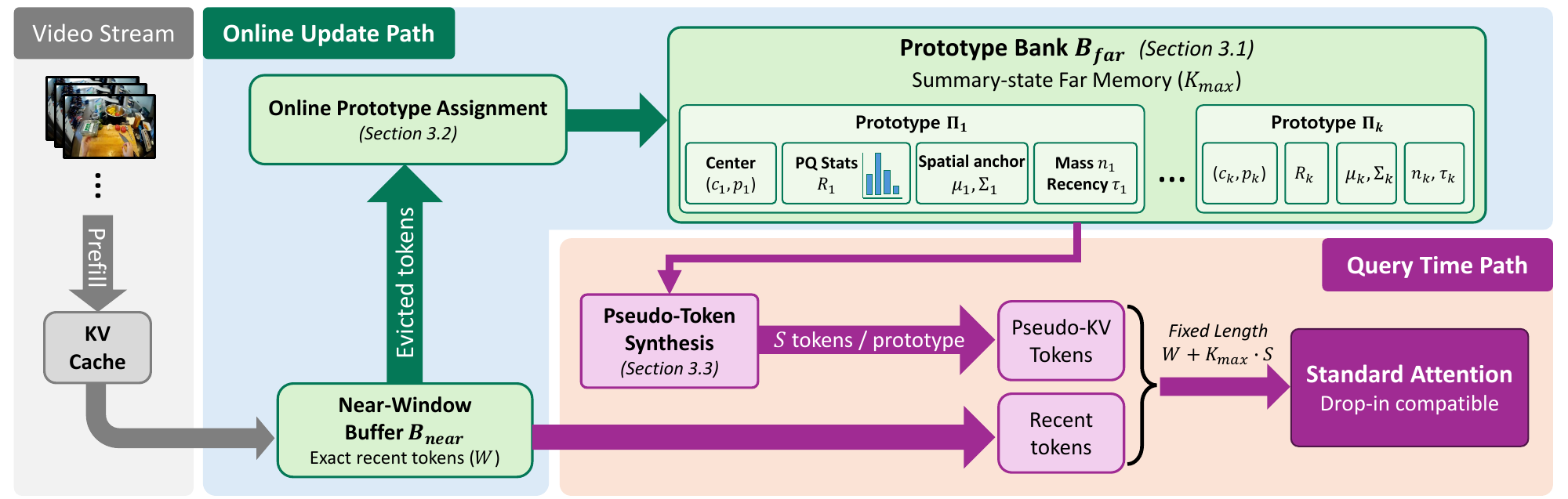} 
    \caption{ProtoKV maintains an exact near-window KV buffer for recent tokens and, as the video stream continues, continuously aggregates evicted history into a fixed-capacity, object-centric prototype bank via online assignment. At query time, it synthesizes a bounded number of pseudo-KV tokens per prototype and concatenates them with the near-window tokens, yielding a fixed-length sequence that can be consumed by standard attention without architectural changes.}
    \label{fig:overview}
\end{figure*}

\subsection{Query Delay and Post-Evidence Update Pressure}
We now formalize query delay as the key stressor in SVU. A decisive cue may appear briefly, yet the system must continue ingesting frames and updating a bounded state until the query arrives. During this interval, the cue can be either explicitly removed or gradually degraded by repeated eviction/compression decisions under a fixed footprint. Fig. \ref{fig:delay} depicts this setting, where evidence occurs at time $t_e$ and the query arrives later at $t_q$ after many intervening updates. In an idealized view, for each query $q$, let $t_e$ denote the last-seen timestamp of the decisive evidence, and $t_q$ denote the query arrival time. We define the query delay as $\Delta_q = t_q - t_e$. Larger $\Delta_q$ means the system must preserve query-relevant information across more intervening streaming updates. 

In practice, SVU benchmarks \cite{rvs, ovobench, streamingbench} typically do not provide per-query evidence timestamps $t_e$. Thus, our delayed-query evaluation uses the dataset-defined query timestamp $t_0$ as a reference and shifts the query to $t_q=t_0+\Delta$ (used as an empirical proxy for $\Delta_q$), while restricting to retrospective query types with invariant ground-truth.

Importantly, $\Delta$ is not merely a time gap; it serves as a proxy for the \emph{amount of post-evidence update pressure} accumulated before answering. During this interval, the system must repeatedly (i) encode incoming frames and (ii) update its bounded internal state. Because these updates are query-agnostic, the system must commit to maintenance decisions before knowing which cues will later be queried, and thus cannot selectively preserve only query-useful evidence. Under any bounded-memory mechanism, far history undergoes many potentially lossy operations such as eviction, quantization, merging, or summarization, whose errors can compound with continued updates. This perspective naturally induces a \emph{near--far regime}: a recent near window can be kept at high fidelity, but once evidence falls outside that window, the model must rely on a more compressed far representation. Accordingly, we often observe a sharp degradation around the near$\rightarrow$far transition, followed by a slower decline as $\Delta$ increases further.

A common strategy for bounded memory is \emph{token retention}, which preserves far history by keeping selected token-level KV instances \cite{h20,snapkv,infinipotv}. However, under large $\Delta$, token retention faces intrinsic challenges. First, retained tokens must compete for a fixed capacity over a long stream; early retention choices can become suboptimal as new content arrives, forcing repeated replacement and yielding compounded information loss. Second, retaining token instances does not directly control whether the remaining set stays \emph{query-usable} after many updates: the mechanism may preserve many redundant or weakly relevant instances while dropping rare but decisive cues, causing abrupt failures once the evidence leaves the near window. These observations motivate representing far history as a \emph{fixed-capacity summary state} rather than as a set of retained token instances, aiming for more graceful degradation as evidence becomes increasingly distant.

\section{ProtoKV}
We present \textbf{ProtoKV}, a constant-footprint KV memory for streaming video LLMs under arbitrary query arrivals. The central challenge is \textit{post-evidence update pressure}: after decisive evidence appears, the system must absorb many subsequent updates before the query arrives, and token-instance retention can become increasingly brittle under a fixed budget as this delay grows. 

ProtoKV maintains a bounded online state with (i)~a footprint independent of stream length, (ii)~query-agnostic online maintenance, and (iii)~drop-in query-time consumption via standard attention. It uses a \textit{two-tier design}: an exact near-window KV cache of length $W$, and a token-instance-free far summary state implemented as a prototype bank of capacity $K_{\max}$. As tokens leave the near window, they are absorbed online into the bank via a \textit{continuity-aware assignment rule}, allowing far memory to accumulate evidence rather than repeatedly overwriting it.

At query time, each prototype is exposed as $S$ pseudo-KV tokens synthesized from its centers and lightweight residual statistics, with mass-aware weighting to reflect token multiplicity. Concatenating near exact KV with far pseudo tokens yields a bounded context length $L_{context} = W + K_{\max} \cdot S$, keeping query-time cost bounded even when decisive evidence is primarily supported by far history. Unlike token-retention schemes that must decide which instances to keep under a fixed budget, we maintain a persistent summary state whose updates refine rather than replace evidence.

\subsection{State representation and prototype bank}
\textbf{Near exact state.} 
We maintain a ring buffer $B_{\text{near}}$ that stores the most recent $W$ tokens' exact KV pairs. This near window preserves fine-grained, temporally local evidence without approximation, serving as an exact anchor when relevant evidence lies within the recent horizon. 

\textbf{Far summary state.} 
For older history, ProtoKV maintains a fixed-capacity prototype bank $B_{\text{far}} = \{ \Pi_k \}_{k=1}^{K_{\max}}$, where each entry $\Pi_k$ is an aggregated summary of many evicted tokens and stores no per-token KV pairs for far history \cite{compressivetransformer}. Each prototype stores only constant-size sufficient statistics: representative centers $(c_k, p_k)$, a mass $n_k$, residual statistics $R_k$, and optional spatiotemporal metadata $(\mu_k, \Sigma_k, \tau_k)$ used for prototype assignment. Consequently, far-memory cost scales with $K_{\max}$\footnote{We keep $K_{max}$ fixed (independent of stream length) under each memory budget; thus update-time complexity is constant with respect to stream length.}, not with the number of processed tokens. Large $\Delta$ stresses far memory primarily via post-evidence update pressure, not mere ‘oldness’. Persistent prototype-level state reduces churn (overwrite/switching), allowing sparse cues to remain represented across many subsequent updates.

\textbf{Initialization.}
As long as empty prototype slots remain, evicted tokens are routed to a new slot rather than absorbed into an active prototype. This prevents premature merging of semantically distinct events under a partially-filled bank and ensures the full $K_{max}$ capacity is utilized as the stream progresses.
Empty prototype slots are initialized upon first assignment by setting the centers from the incoming token and resetting the associated statistics.

\subsection{Online update: eviction, assignment, and prototype maintenance}
\label{online_update}
As the stream progresses, the KV cache of each new token is appended to $B_{\text{near}}$. When $|B_{\text{near}}| > W$, the KV pair of the oldest token is evicted and absorbed into $B_{\text{far}}$. A key difficulty is that far memory is updated continuously after a decisive cue occurs; without a continuity-aware rule, evicted tokens from the same underlying object/track can be scattered across prototypes, causing prototype switching and fragmentation under sustained update pressure. To address this, we assign evicted tokens using a continuity-aware objective that jointly considers key similarity, spatial consistency, and recency, and then update the selected prototype as a persistent summary state.

\textbf{Prototype assignment.} 
For an evicted token $i$ and KV cache ($K_i, V_i)$, let $s_i = (x_i, y_i)$ denote its normalized 2D spatial coordinate in the frame. We assign it to the prototype minimizing the online cost:
\begin{equation}
\begin{aligned}
k^{\star} = \arg \min_{k} \Big[ &-\cos(K_i, c_k) + \lambda_{\text{sp}} d_{\text{Mah}}(s_i; \mu_k, \Sigma_k) \\
&+ \lambda_{\mathrm{idle}} \mathbf{1}[idle(\tau_k)] \Big]
\end{aligned}
\end{equation}

The objective combines three terms. 
(i) \textit{Key-space similarity}: $-\cos(K_i, c_k)$ groups tokens whose keys are close to the prototype's key center $c_k$. We use keys only for assignment because, in dot-product attention, retrieval weights depend on $q^\top k$ while values affect only the retrieved content after weighting \cite{transformer}; values are typically more content-variant and thus less reliable for stable online grouping. 
(ii) \textit{Spatial continuity}: $\lambda_{\text{sp}} d_{\text{Mah}}(s_i; \mu_k, \Sigma_k)$ enforces track-consistent association under prototype $k$'s running location distribution $(\mu_k, \Sigma_k)$, acting as a prototype-specific elliptical gate that reduces spurious switching when multiple similar objects co-occur \cite{deepsort}. 
(iii) \textit{Staleness control}: $\lambda_{\mathrm{idle}} \mathbf{1}[idle(\tau_k)]$ discourages matching to prototypes that have not been updated recently, where $idle(\tau_k)$ denotes the predicate $t-\tau_k > T_{idle}$ with $t$ being the current stream time.
Once the bank is populated, continuity-aware assignment reduces prototype switching and fragmentation even when many updates occur after the decisive cue, preventing evidence from being scattered and diluted across prototypes.

\begin{table*}[t]
\caption{Main benchmark performance on SVU benchmarks.
We use token budget $|M|$ = 24k tokens for RVS-Ego and RVS-Movie, and $|M|$ = 4k tokens for OVO-Bench and StreamingBench.
For StreamingBench, we evaluate only Real-time visual understanding queries. Score columns are on a 0-5 scale.}
\centering
\small
\setlength{\tabcolsep}{6pt}
\begin{tabular}{l l cc cc c c}
\toprule
Backbone & Method &
\multicolumn{2}{c}{RVS-Ego} &
\multicolumn{2}{c}{RVS-Movie} &
OVO-Bench &
StreamingBench \\
\cmidrule(lr){3-4}\cmidrule(lr){5-6}\cmidrule(lr){7-7}\cmidrule(lr){8-8}
& & Acc. (\%) & Score & Acc. (\%) & Score & Acc. (\%) & Acc. (\%) \\
\midrule

\multirow{3}{*}{LLaVA-OV-7B}
& SWA & 55.7 & 3.30 & 50.8 & 3.40 & 53.5 & 72.9 \\
& InfiniPot-V    & 57.8 & 3.50 & 51.4 & 3.50 & 54.2 & 75.2 \\
& \textbf{ProtoKV} & \textbf{58.6} & \textbf{3.63} & \textbf{52.1} & \textbf{3.59} & \textbf{54.8} & \textbf{76.3} \\
\midrule

\multirow{3}{*}{Qwen2.5-VL-7B}
& SWA & 58.4 & 3.60 & 52.0 & 3.55 & 53.2 & 75.9 \\
& InfiniPot-V    & 58.9 & 3.75 & 52.3 & 3.69 & 53.6 & 76.4 \\
& \textbf{ProtoKV} & \textbf{59.4} & \textbf{3.84} & \textbf{53.1} & \textbf{3.77} & \textbf{54.4} & \textbf{77.3} \\
\bottomrule
\end{tabular}
\vspace{2pt}
\label{tab:main_benchmark}
\end{table*}

\textbf{Prototype update.} 
After assigning token $i$ to $k^{\star}$, we update the selected entry by (i) updating centers $(c_{k^{\star}}, p_{k^{\star}})$ (e.g., via EMA), (ii) incrementing mass $n_{k^{\star}}$, and (iii) updating residual statistics $R_{k^{\star}}$ using the token's key/value residuals. We also update the spatial state $(\mu_{k^{\star}}, \Sigma_{k^{\star}})$ with $s_i$ and refresh recency metadata $\tau_{k^{\star}}$. This assignment is lightweight, online, and query-agnostic and is orthogonal to our query-time pseudo-token interface; alternative online association rules can be substituted without changing the pseudo-token mechanism (see \cref{alg:streaming} for details).

\textbf{Capacity management.}
Beyond per-token updates, ProtoKV applies lightweight maintenance to keep the bank stable over long streams (\cref{alg:maintenance}). Idle prototypes whose recency $\tau_k$ has not been refreshed for $T_{idle}$ steps undergo mass decay, redundant prototypes whose centers are sufficiently close are merged, and slots whose mass has decayed to zero are recycled by reinitializing from a recent token in $B_near$. This preferentially preserves frequently supported prototypes while keeping the policy fixed across experiments. Since decisive cues are first retained exactly in the near window and then absorbed via continuity-aware assignment, prototypes that continue receiving consistent support quickly accumulate mass and are unlikely to be recycled immediately.

\subsection{Residual Statistics and Pseudo-Token Interface}
Prototype centers $(c_k, p_k)$ are a strong constant-memory summary, but center-only aggregation is fundamentally lossy: as more tokens are absorbed, distinctive cues are averaged out. This matters most under large query delay $\Delta$, where evidence must remain usable after many post-evidence updates: this is precisely the regime where decisive evidence becomes temporally distant and far-memory quality governs whether it remains usable.

\textbf{Residual statistics.} 
For a token $i$ assigned to prototype $k$, we define residuals: $r_i^K = K_i - c_k, \quad r_i^V = V_i - p_k$.
The key point is not to reconstruct individual tokens, but to retain intra-prototype diversity that a single center cannot represent. We implement this with Product Quantization \cite{pq}-based streaming histogram statistics: it offers a fixed-budget summary that (i) can be updated online without storing per-token codes, and (ii) captures multi-pattern residual structure around the center, reducing the chance that decisive cues are washed out as updates accumulate.

\textbf{Pseudo tokens.} 
Residual statistics are useful only if they can be consumed at query time without changing the model.
We expose far memory as a fixed set of pseudo tokens: synthetic KV-cache entries that are directly consumable by standard attention.
Injecting a small set of auxiliary key/value vectors as a lightweight interface to attention has been explored in parameter-efficient adaptation \cite{prefixtuning}.
Unlike learned prefixes, our pseudo tokens are synthesized online in a training-free manner from the prototype bank statistics.
ProtoKV converts each prototype into $S$ pseudo tokens by combining its center with representative residual patterns implied by the stored statistics $R_k$.
These pseudo-tokens are concatenated with the near-window exact KV and processed by standard attention unchanged, resulting in a fixed-length query context.
This design keeps the per-query context size predictable, controlling query-time latency and attention cost even when decisive evidence lies in far history.
For positional encoding, all $S$ pseudo-tokens synthesized from prototype $k$ share the same rotary position $\tau_k$, the original stream position of the most recently absorbed source token, while near-window tokens retain their original positions.
This preserves prototype-level temporal recency without per-pseudo-token position tracking and keeps the interface fully compatible with standard attention (see \cref{sec:rope_appendix} for a comparison against alternative anchor choices).

\textbf{Mass-aware bias.} 
A prototype summarizes $n_k$ absorbed tokens; without accounting for
multiplicity, far memory can underweight frequently observed evidence
relative to rare prototypes. We therefore apply a simple per-prototype
log-mass bias $b_k = \log n_k$ to the attention logits. For each
attention head, the logit of pseudo-token $s$ decoded from prototype
$k$ becomes
\begin{equation}
  \ell_{k,s}
  = \frac{q^{\top} \tilde{k}_{k,s}}{\sqrt{d}} + b_k,
\end{equation}
where $b_k$ is shared across all $S$ pseudo-tokens of prototype $k$
and applied before the softmax. This approximates the effect of having
$n_k$ duplicated tokens without instantiating them, so that a
prototype's influence in attention is proportional to its accumulated
evidence.
This log-count additive correction is analogous to prior-based logit adjustments under softmax normalization \cite{longtail1, longtail2}.

\section{Experiments}

\subsection{Experimental Setup}

\begin{figure*}[t]
  \centering
  \includegraphics[width=0.8\textwidth]{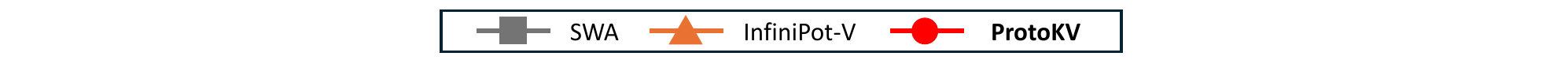} 
    \vspace{10pt}
  \begin{subfigure}[t]{0.34\textwidth}
    \centering
    \includegraphics[width=\linewidth]{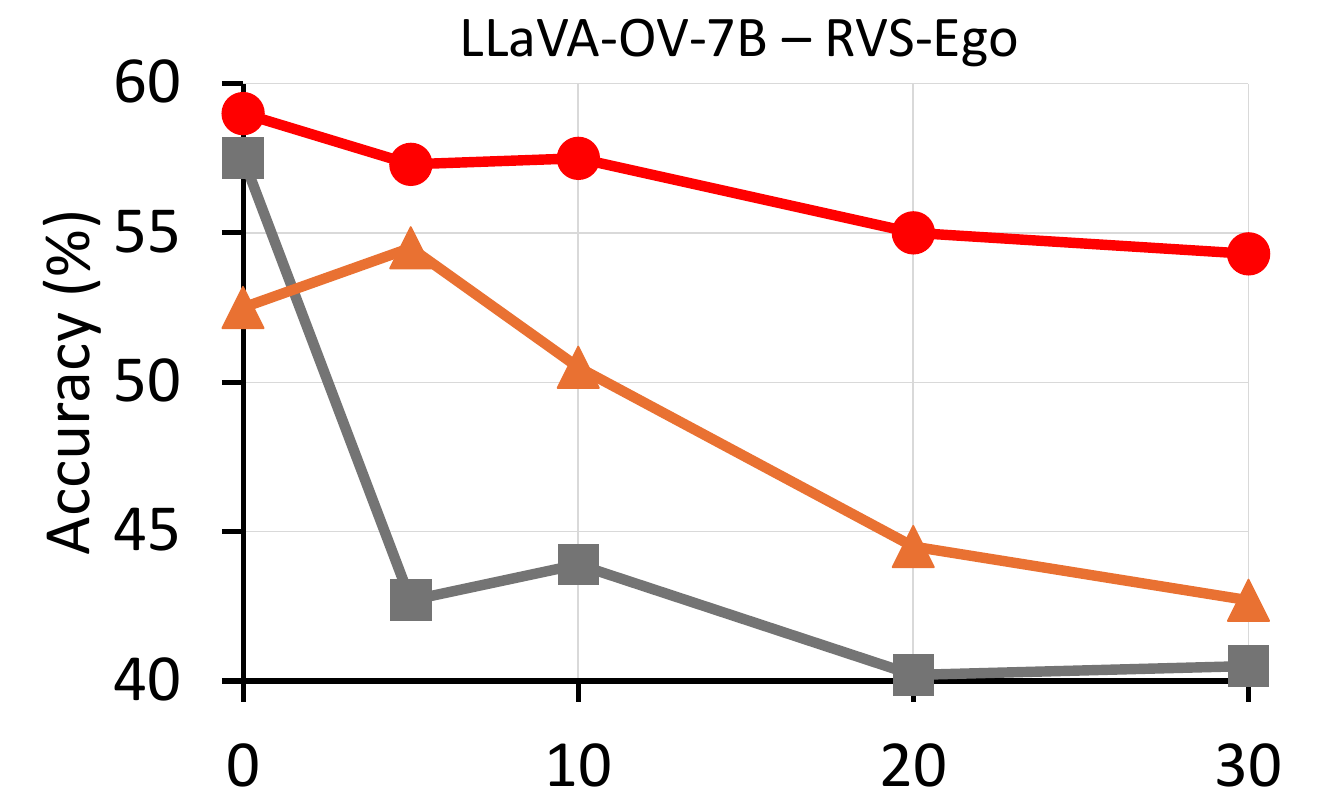}
    \label{fig:delay_ego_llava}
  \end{subfigure}
  \hfill
  \begin{subfigure}[t]{0.312\textwidth}
    \centering
    \includegraphics[width=\linewidth]{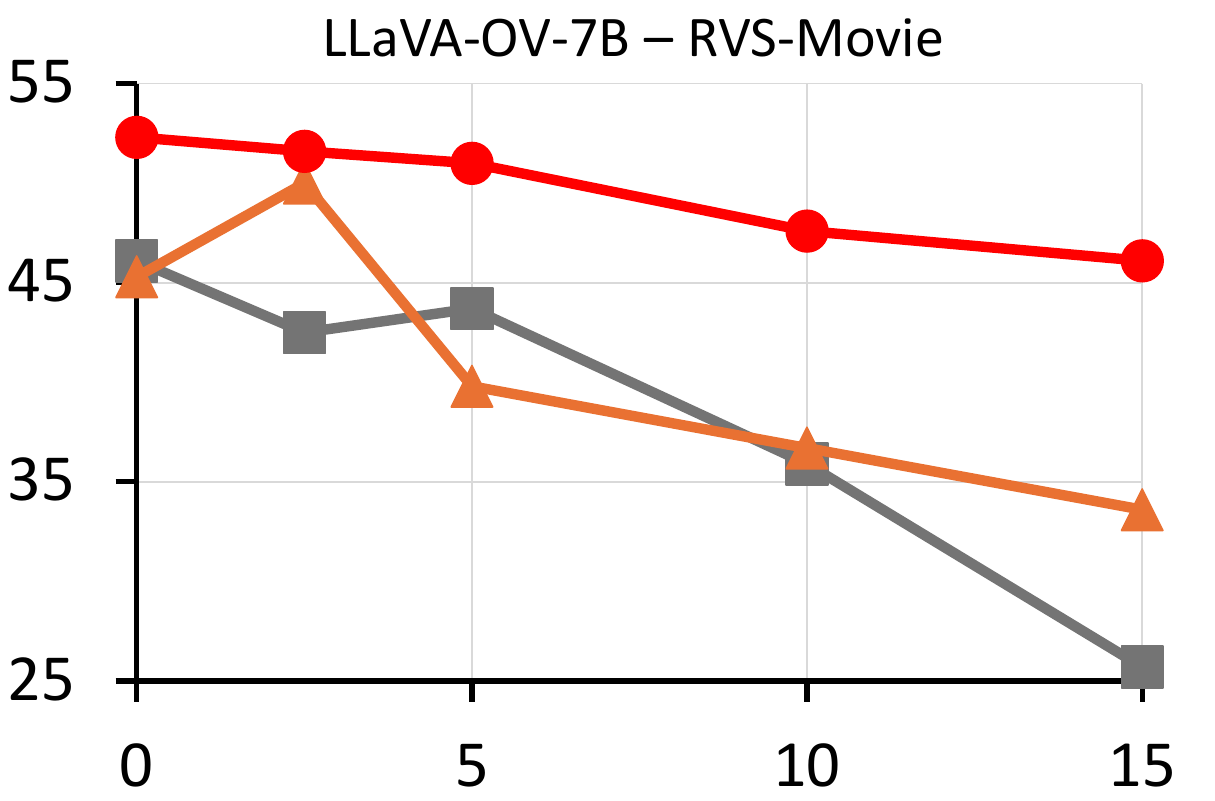}
    \label{fig:delay_movie_llava}
  \end{subfigure}
  \hfill
  \begin{subfigure}[t]{0.312\textwidth}
    \centering
    \includegraphics[width=\linewidth]{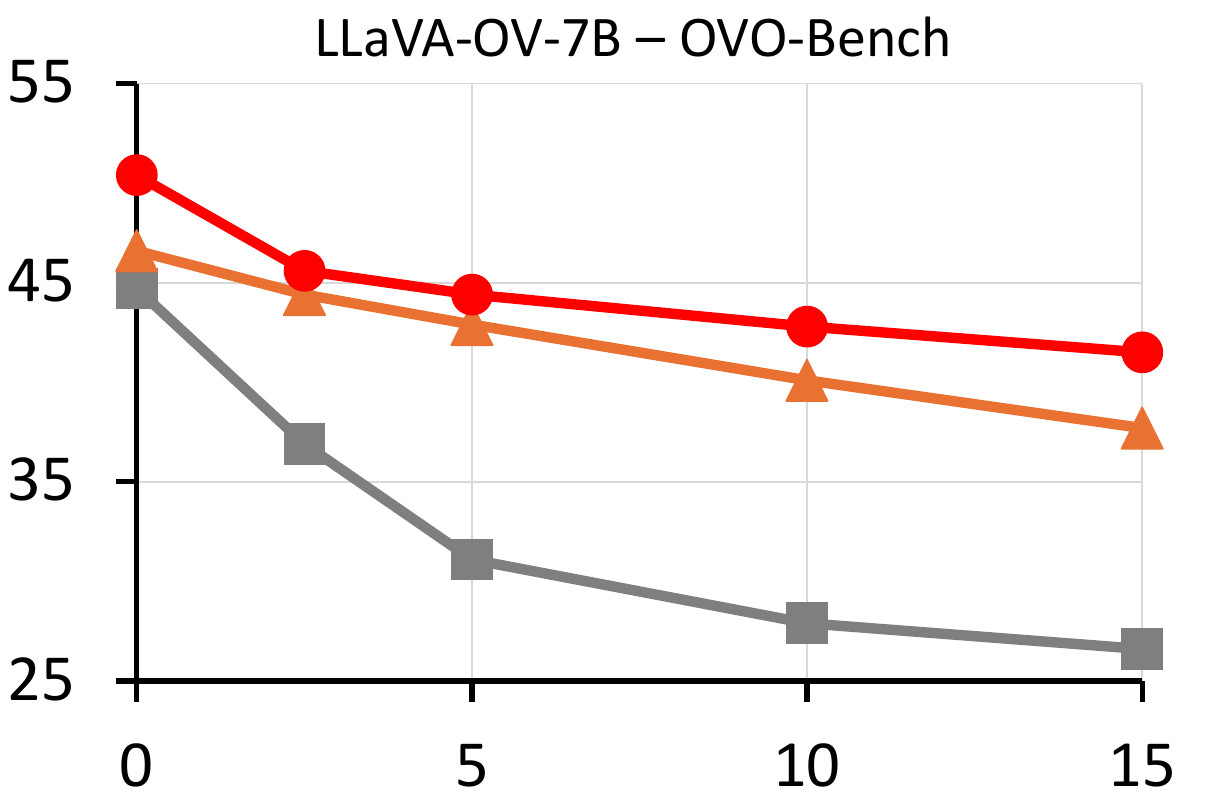}
    \label{fig:delay_ovo_llava}
  \end{subfigure}

    \vspace{-15pt}

  \begin{subfigure}[t]{0.34\textwidth}
    \centering
    \includegraphics[width=\linewidth]{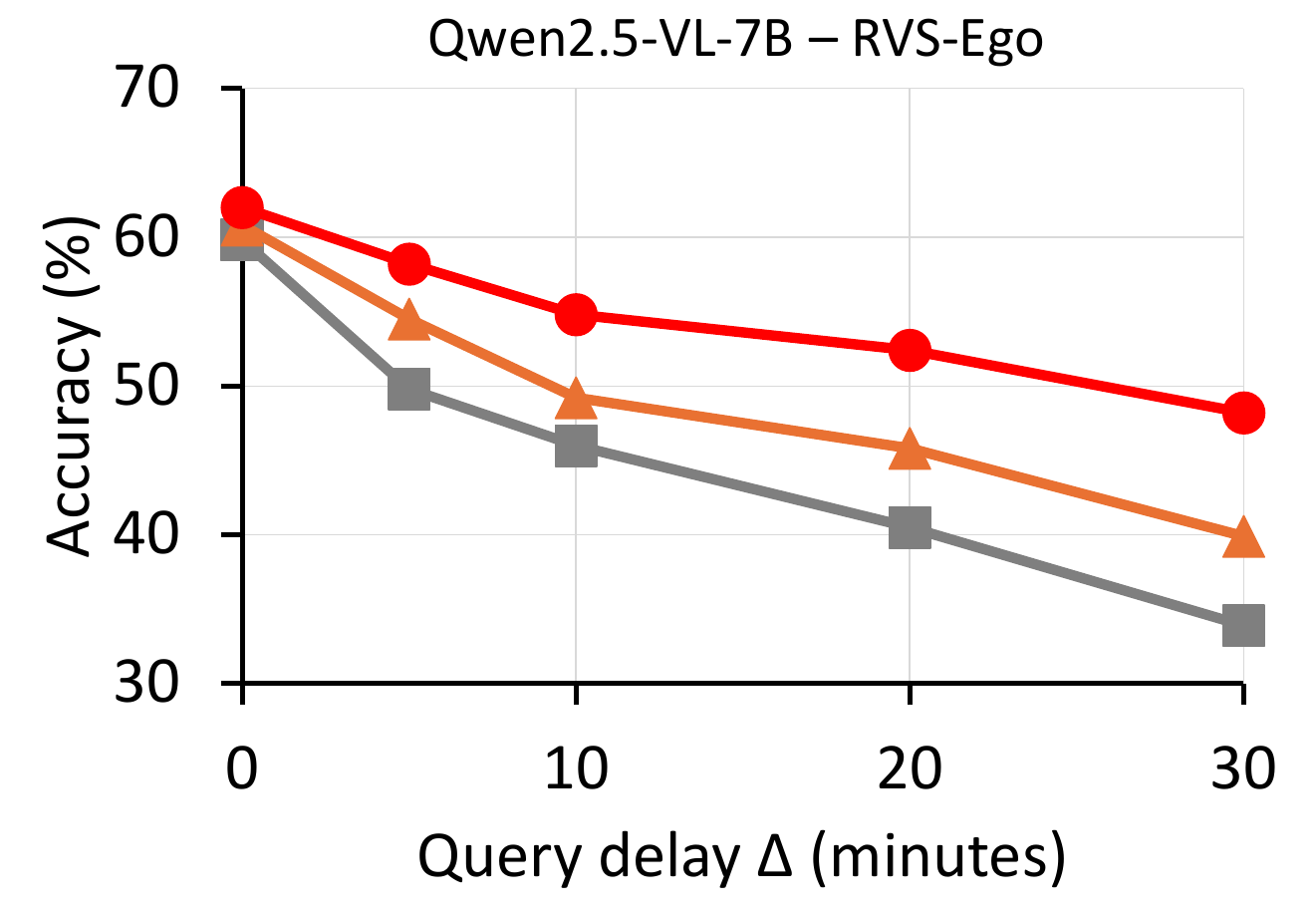}
    \label{fig:delay_ego_qwen}
  \end{subfigure}
  \hfill
  \begin{subfigure}[t]{0.312\textwidth}
    \centering
    \includegraphics[width=\linewidth]{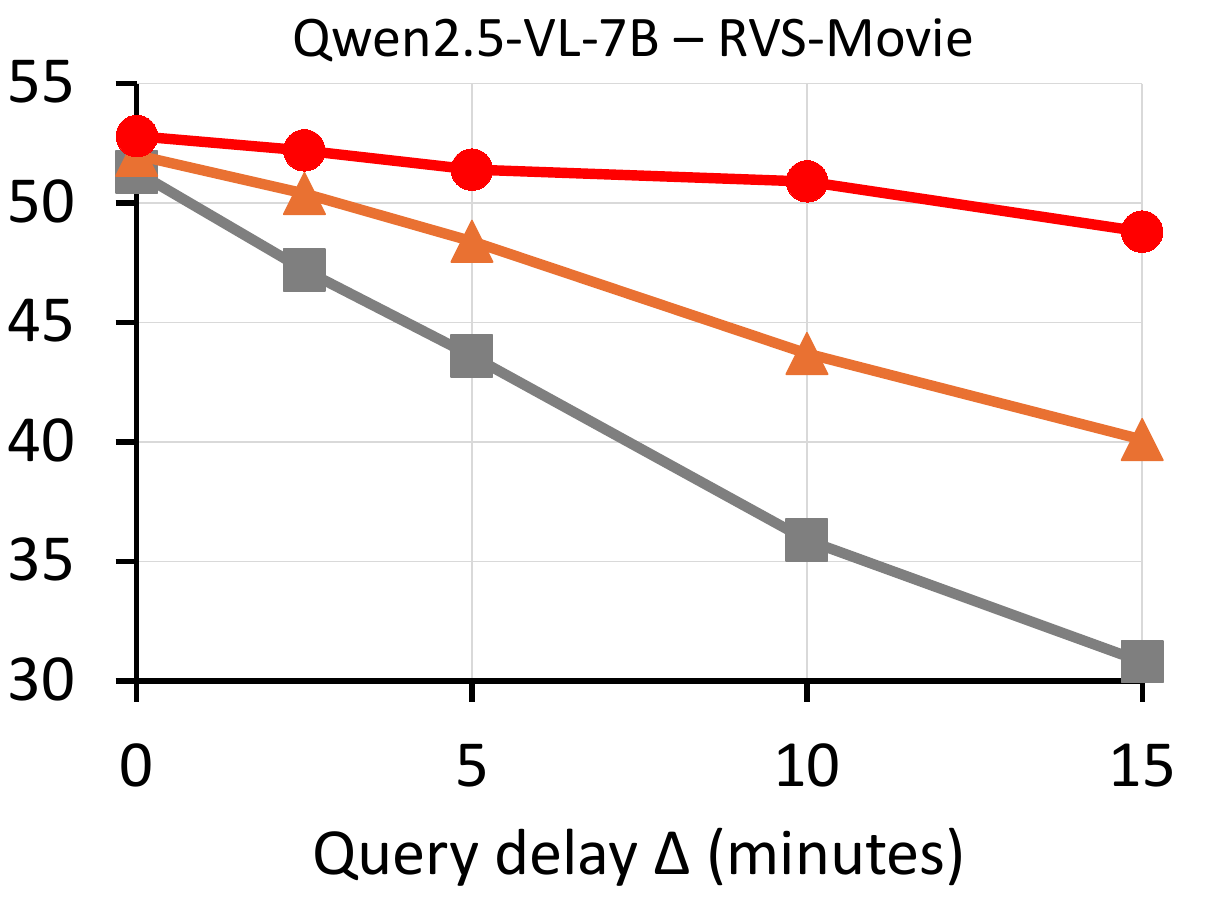}
    \label{fig:delay_movie_qwen}
  \end{subfigure}
  \hfill
  \begin{subfigure}[t]{0.312\textwidth}
    \centering
    \includegraphics[width=\linewidth]{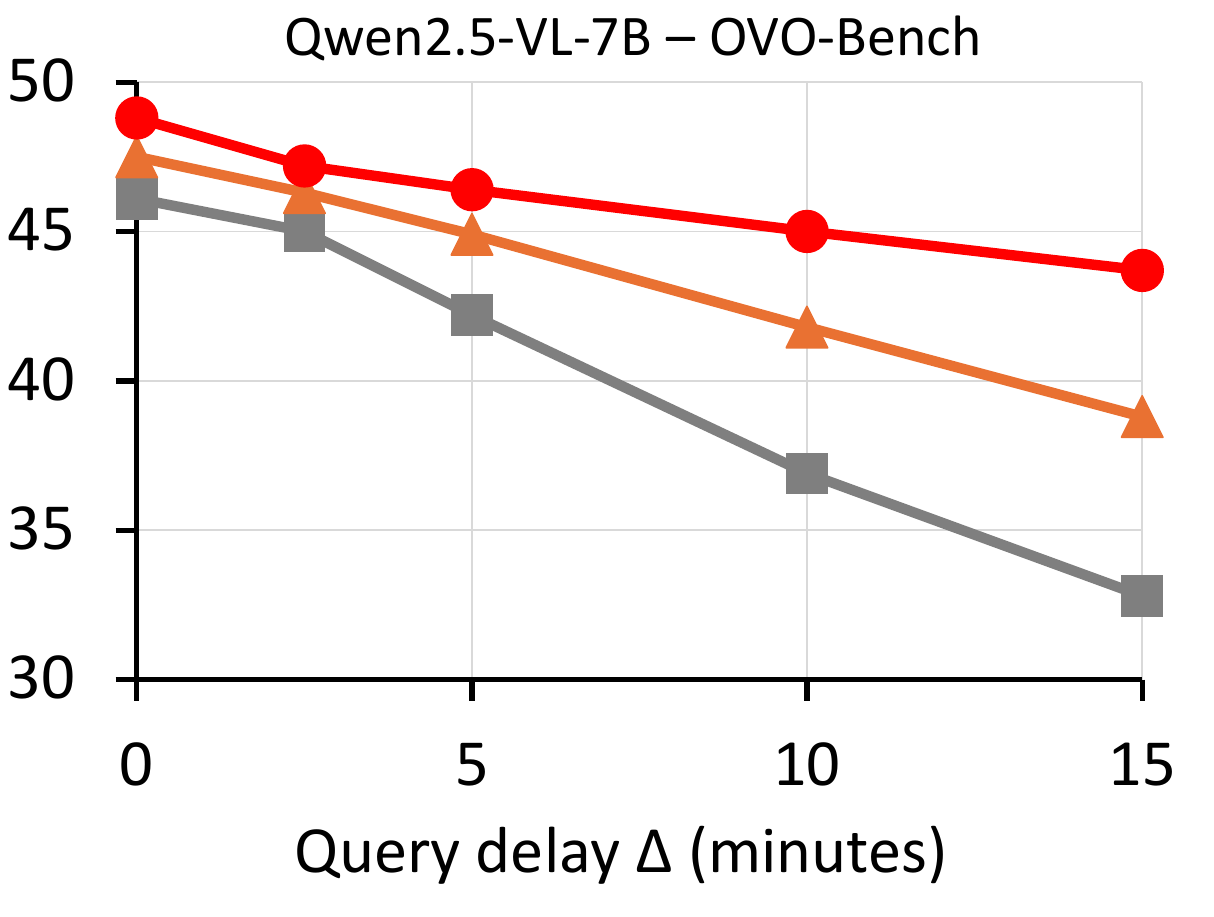}
    \label{fig:delay_ovo_qwen}
  \end{subfigure}

  \vspace{-4pt}
  \caption{Query-delay robustness under matched token budgets in the online SVU setting. We apply a delayed-query sweep by setting $t_q = t_0 + \Delta$ (minutes) and retain only retrospective query types whose answers are invariant to shifting $t_q$. For fair comparisons across $\Delta$, we evaluate on the intersection subset that is valid for all delay values. We use token budget $|M|$ = 24k tokens for RVS-Ego and RVS-Movie, and $|M|$ = 4k tokens for OVO-Bench.}
  \label{fig:delay_sweep}
  \vspace{-6pt}
\end{figure*}

\textbf{Benchmarks.} We evaluate on four SVU benchmarks \cite{rvs, ovobench, streamingbench} that provide a dataset-defined query timestamp for each example and task-type annotations that allow principled grouping of queries. We follow the official evaluation split for each dataset. These annotations enable controlled delayed-query sweeps and rule-based filtering of query types without manual curation. For RVS-Ego and RVS-Movie \cite{rvs}, we follow the benchmark's official evaluation pipeline and use \texttt{gpt-3.5-turbo-0125} to compute the answer score on a 0–5 scale, where higher is better.

\textbf{Delayed-Query Protocol and Valid-Query Selection.} Ideally, one would evaluate query delay relative to the last-seen evidence time $t_e$ by setting $t_q = t_e + \Delta_e$. Since the evaluated SVU benchmarks do not provide $t_e$ per query, we instead use the dataset-defined query timestamp $t_0$ as a reference and sweep delayed queries by setting $t_q = t_0 + \Delta$.

To keep ground-truth unchanged under this shift, we apply a rule-based valid-query filter and evaluate only retrospective query types whose answers depend on past video content rather than on events occurring after $t_0$. For query types that are inherently time-dependent (e.g., future-facing or real-time queries), we do not include them in query-delay robustness experiments. See Appendix~\ref{sec:dataset_selection} for further details.

For fair comparisons across delay values, we evaluate on the intersection subset of queries that remain valid for all $\Delta$ in the sweep. Concretely, for each example with video length $L$, we keep it only if $t_0 + \Delta \leq L$ holds for every evaluated $\Delta$, ensuring that all delay points are computed on the same set of queries. All $\Delta$ values reported in the experiments are measured in minutes.

\textbf{Baselines.}
We restrict our comparisons to methods that match our deployment constraints: (i) bounded KV memory, (ii) training-free evaluation (no additional fine-tuning), and (iii) query-agnostic online updates, where the memory state is continuously maintained as frames arrive and queries are answered immediately at tq without query-conditioned retrieval or re-encoding. Under these criteria, we include two representative designs. First, Sliding-window attention (SWA) serves as the most naive bounded-memory baseline: it guarantees constant memory simply by retaining only the most recent window of KV pairs and truncating all earlier history. Second, query-agnostic continual KV token retention \cite{infinipotv} maintains a fixed KV budget under a hard cap by selecting, merging, or evicting token instances online. We instantiate ProtoKV on two 7B-scale vision-language backbones, LLaVA-OV-7B \cite{llava-ov} and Qwen2.5-VL-7B \cite{qwen25} with identical pretrained weights and no fine-tuning unless noted.

\textbf{Budget Matching and Efficiency Measurement.} 
We compare methods under a common query-time context budget by controlling the number of visual tokens that each method allows the model to attend to at query time. We then report the resulting measured peak GPU memory and latency metrics to verify that the compared settings fall within the intended budget regime. Query-time latency is reported separately in Fig. \ref{fig:ttft}.

\subsection{Main Benchmark - SVU}
We first evaluate ProtoKV on standard SVU benchmarks under matched peak GPU memory; query-time latency is validated separately in Fig. \ref{fig:ttft} under the same budgeted setting. Table \ref{tab:main_benchmark} reports main benchmark accuracy and scores for two backbones and compares ProtoKV against SWA and a token-retention baseline.

ProtoKV achieves the best overall performance across the RVS benchmarks on both backbones, consistently outperforming SWA and the token-retention baseline. ProtoKV likewise yields higher accuracy on OVO-Bench and StreamingBench under the same KV cache memory budget. Overall, these results indicate that ProtoKV’s summary-state far memory makes more effective use of limited memory than window truncation or token-instance retention, especially when decisive evidence must be preserved over long update horizons.

\subsection{Query-Delay Robustness}
Fig. \ref{fig:delay_sweep} reports accuracy as a function of query delay $\Delta$. Across all settings, performance decreases as $\Delta$ grows, but the degradation profile differs sharply by memory mechanism. SWA exhibits an abrupt drop once the delay pushes the relevant evidence beyond the near window, reflecting that evidence is irrecoverably lost after truncation. InfiniPot-V degrades more gradually, yet still shows a consistent downward trend as post-evidence updates accumulate under a fixed budget. In contrast, ProtoKV remains substantially more stable over long delays: the slope of degradation is noticeably flatter, and the relative gains over both SWA and token retention increase with $\Delta$—precisely in the regime where SVU must rely on far memory rather than near-window access.

\begin{table}[t]
\caption{Component ablations under matched peak GPU memory ($|M|$ = 24k, model: LLaVA-OV-7B, dataset: RVS-Ego). Evaluated on the same intersection subset used in \cref{fig:delay_sweep}. }
\centering
\small
\begin{tabularx}{0.9\columnwidth}{lcc}
\toprule
Setting & \makecell{Acc.@$\Delta{=}0$ \\ (\%)} & \makecell{Acc.@ $\Delta{=}30$ \\ (\%)}  \\
\midrule
\textbf{ProtoKV (full)} & 59.0 & \textbf{54.3} (\textbf{-4.7}) \\
\midrule
\multicolumn{3}{l}{\emph{Memory tier}}\\
Near only (w/o far)      & 57.5 & 40.5 (-17.0)  \\
Far only (prototype)     & 54.5 & 49.2 (-5.3)  \\
\midrule
\multicolumn{3}{l}{\emph{Far-memory expression}}\\
w/o residual stats       & 58.5 & 51.0 (-7.5)  \\
w/o mass bias ($b_k{=}0$)& 58.3 & 51.0 (-7.3)  \\
\midrule
\multicolumn{3}{l}{\emph{Assignment}}\\
w/o continuity ($\lambda_{sp}{=}0$) & 56.2 & 47.5 (-8.7)\\
\bottomrule
\end{tabularx}
\label{tab:ablation}
\end{table}

At the largest evaluated delays, across both backbones, ProtoKV yields up to +20.4pp over SWA and +12.5pp over token retention on RVS-Movie, with comparable long-delay improvements on RVS-Ego and OVO-Bench. This behavior supports our central interpretation of $\Delta$ as post-evidence update pressure, not merely “older evidence”: larger $\Delta$ implies the decisive cue must survive more overwrite/eviction/merge events under the same capacity. These results indicate that ProtoKV’s fixed-capacity summary-state updates mitigate brittleness under long delays while preserving bounded query-time cost.

\subsection{Ablation Study}
Table \ref{tab:ablation} summarizes component ablations under matched peak GPU memory. ProtoKV shows a relatively mild drop as query delay increases, while the ablations reveal that this robustness is not due to any single trick but to the overall design. In particular, removing the far-memory summary leads to a pronounced failure at long delays, whereas keeping only the far prototypes preserves delayed-query behavior but sacrifices performance in the near regime. This highlights that ProtoKV’s robustness arises from the intended complementarity between an exact near window for immediate queries and a durable far summary for delayed queries.

Among far-memory components, removing residual statistics or mass-aware weighting worsens delayed-query performance, and disabling continuity-aware association further increases degradation, aligning with the need for both expressive summaries and stable online maintenance to avoid evidence fragmentation under sustained updates. Consistent with this, ProtoKV’s long-delay stability depends jointly on (i) preventing over-averaging in the far summary and (ii) maintaining stable assignment over time under a hard budget.

\begin{table}[t]
\caption{Performance on offline video understanding benchmark (OVU) under fixed memory budgets. * denotes the numbers from official paper. We use Qwen2-VL-7B in this table for alignment with prior OVU baselines.}
\centering
\small
\begin{tabular}{l c cc c}
\toprule
Compression & Budget & VideoMME & MLVU  \\
Method & $|M|$ & Acc. (\%) & Acc. (\%) \\
\midrule
\multirow{2}{*}{DyCoke*} & 3K & 55.3 & 57.5 \\
 & 6K & 59.7 & 60.5 \\
\midrule
\multirow{2}{*}{InfiniPot-V*} & 3K & 61.2 & 67.2 \\
 & 6K & 62.8 & 68.4 \\
\midrule
\multirow{2}{*}{\textbf{ProtoKV}} & 3K & \textbf{62.1} & 67.0  \\
 & 6K & \textbf{63.7} & \textbf{68.9}\\
\bottomrule
\end{tabular}
\label{tab:ovu_bench}
\end{table}

\subsection{Auxiliary Evaluation on Offline Video Understanding}
To complement our SVU evaluation, we additionally evaluate ProtoKV on offline video understanding benchmarks \cite{videomme, mlvu} under the same memory budget, using Qwen2-VL-7B \cite{qwen2_} to align with prior OVU baselines' reported settings. We compare ProtoKV against representative OVU baselines including DyCoke \cite{dycoke}. As shown in Table \ref{tab:ovu_bench}, ProtoKV improves performance on VideoMME under tight budgets and remains competitive on MLVU, staying close to the best baseline across budgets. Overall, these results suggest that ProtoKV can be effective beyond SVU-specific evaluations under memory-constrained offline video understanding.

\subsection{Task-Type Analysis}
To examine where summary-state far memory helps and where it falls short, we
decompose ProtoKV's behavior along MLVU's task taxonomy
(\cref{sec:task_breakdown}). ProtoKV is most reliable on tasks whose
answers depend on localized state cues or relations among visually
distinct events. The clearest weakness appears in tasks that require
separating or counting many visually similar repeated events: under a
fixed prototype capacity, such occurrences are absorbed into the same
prototype and their boundaries are blurred by compression---a
mechanistic trade-off of representing far history as a fixed-capacity
summary rather than as token instances. Representative cases are
provided in \cref{sec:case_studies}.

\subsection{System Validation}
\begin{figure}[t]
  \centering
  
  \begin{subfigure}[]{0.48\columnwidth}
    \centering
    \includegraphics[width=\columnwidth]{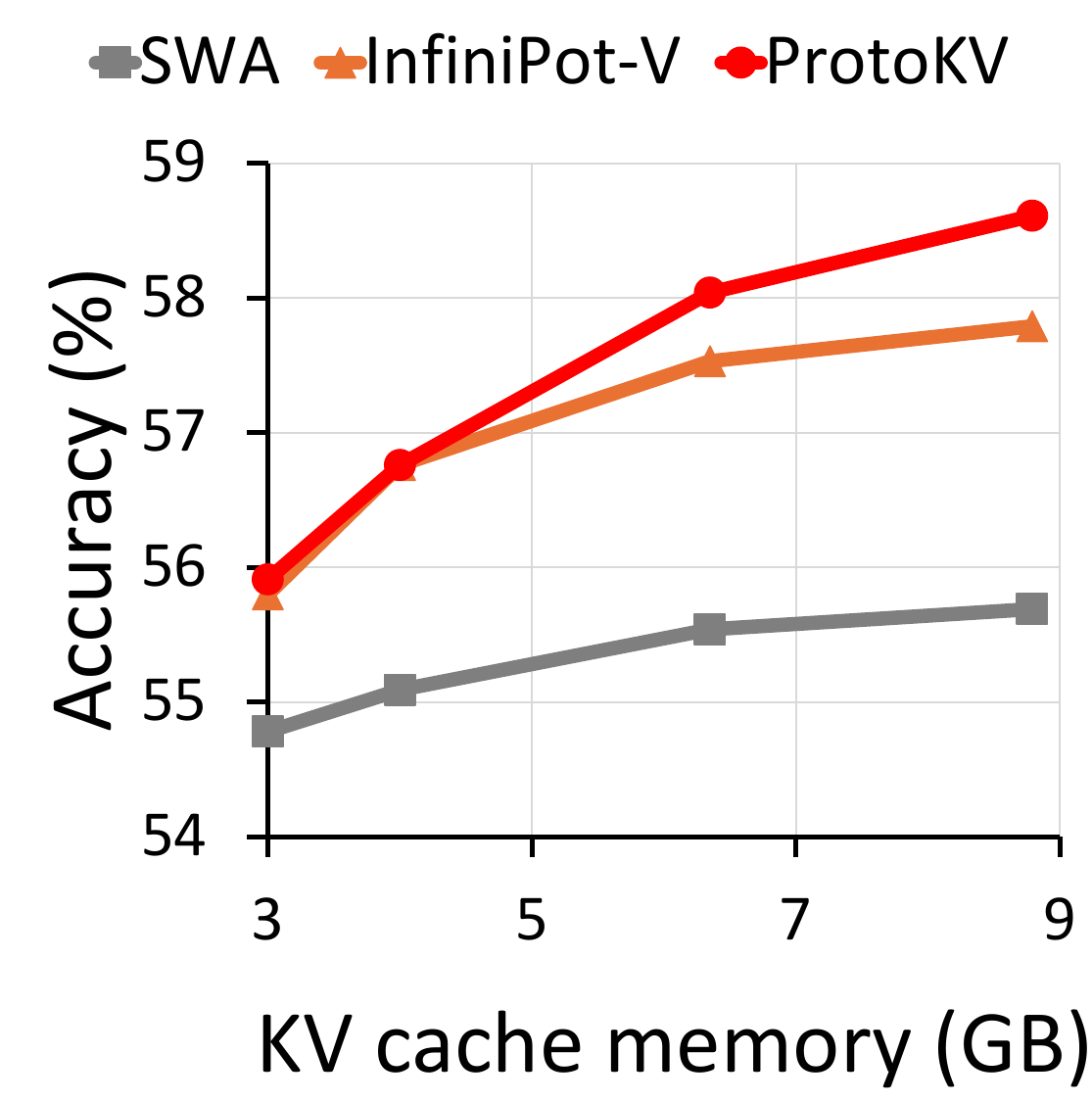}
    \caption{Memory budget}
    \label{fig:memory}
  \end{subfigure}
  \hfill 
  \begin{subfigure}[]{0.48\columnwidth}
    \centering
    \includegraphics[width=\columnwidth]{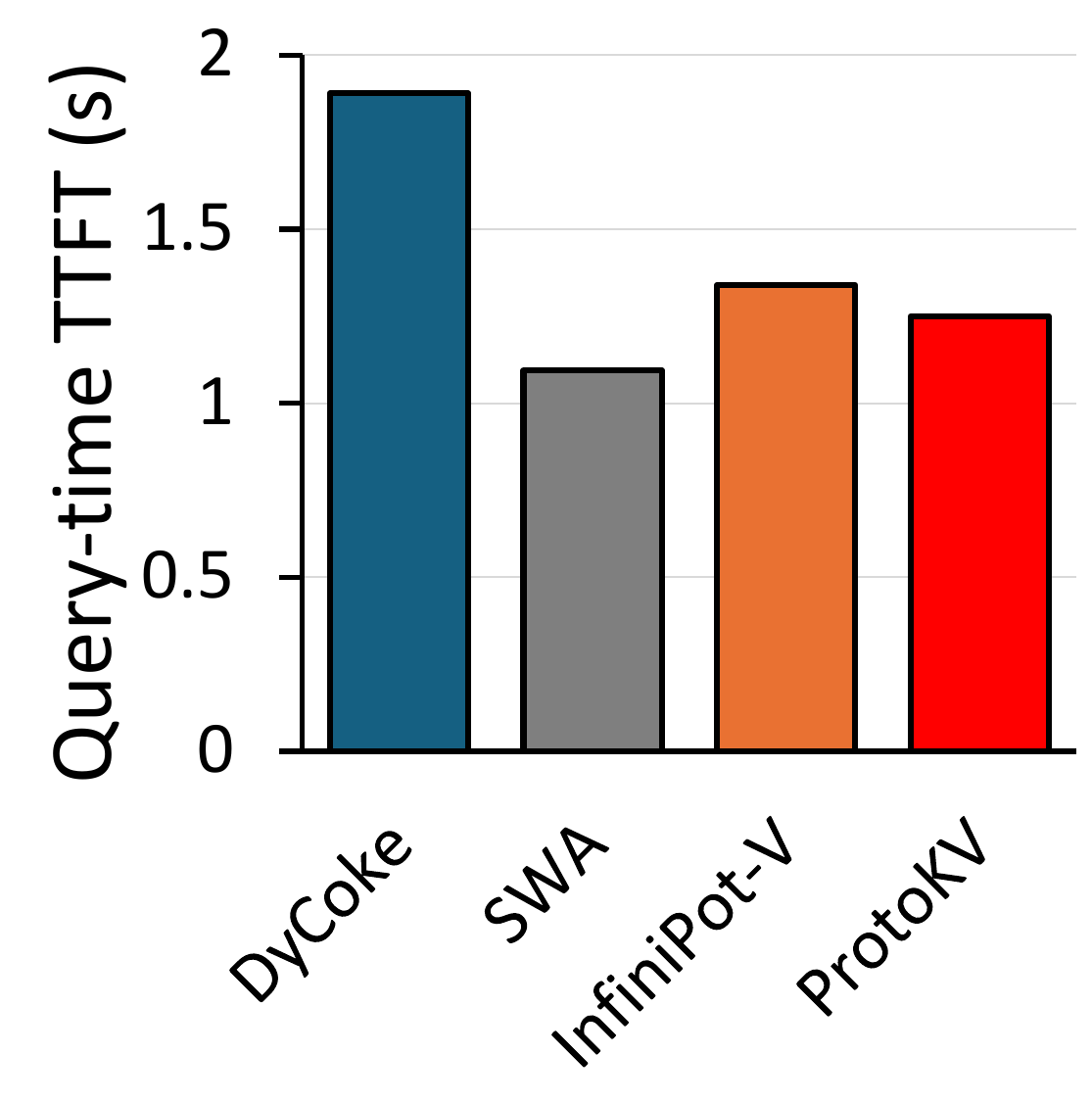}
    \caption{Query-time latency}
    \label{fig:ttft}
  \end{subfigure}
  \caption{Budget sensitivity and query-time TTFT under memory constraints. (a) Accuracy versus memory budget. (b) Query-time TTFT (query arrival → first token) under a matched memory budget. DyCoke is shown as a query-conditioned reference.}
  \label{fig:system}
\end{figure}

Fig. \ref{fig:system} summarizes the system-level budget sensitivity of ProtoKV under memory constraints. Fig. \ref{fig:memory} reports accuracy as we vary the memory budget, showing that ProtoKV consistently delivers the strongest accuracy among the budgeted baselines, with larger budgets translating into more reliable improvements. Fig. \ref{fig:ttft} compares query-time TTFT (time from query arrival to the first generated token) under a matched memory budget. ProtoKV attains competitive query-time TTFT relative to budgeted baselines, and is substantially faster than DyCoke \cite{dycoke}, an OVU-oriented, query-conditioned reference method whose query-time selection introduces additional overhead. Overall, Fig. \ref{fig:system} validates that ProtoKV achieves favorable accuracy–latency trade-offs under fixed memory budgets, aligning with the low-latency requirements of streaming scenarios.

\textbf{Comparison to SVU offloading baseline.}
Table \ref{tab:rekv} compares ProtoKV with ReKV \cite{rekv}, a representative offloading/retrieval-based SVU baseline, to contextualize the latency--accuracy trade-off against an external-memory approach under matched peak GPU memory. While achieving comparable accuracy on RVS-Ego, ProtoKV improves query-time responsiveness, reducing both query-time TTFT and end-to-end latency by 26\%. This indicates that a GPU-resident constant-footprint memory can better satisfy the low-latency requirements of streaming SVU with asynchronous queries and strict response-time budgets.

\textbf{Online update overhead.} 
Beyond query-time TTFT, we measure ProtoKV's online cost during stream
ingestion on Qwen2.5-VL-7B (RTX 5090, RVS-Ego), and find a per-frame
update overhead of $33.8$\,ms, about $30.8\%$ of total per-frame
processing time (\cref{sec:overhead_appendix}). This cost is incurred
during ingestion rather than at query time and can be partially
overlapped with inter-frame gaps in sampled-frame SVU pipelines, so it
does not directly affect query-time TTFT.

\begin{table}[t]
\caption{Comparison to an SVU offloading/retrieval baseline on RVS-Ego under matched peak GPU memory with LLaVA-OV-7B. TTFT (query arrival $\rightarrow$ first token) and E2E latency (query arrival $\rightarrow$ completion) are measured.
}
\centering
\small
\begin{tabular}{l c c c c}
\toprule
Compression & Acc. & Peak GPU & TTFT & E2E Lat. \\
Method & (\%) & (GB)  & (s) & (s)\\
\midrule
ReKV    & 58.8  & 23.0 & 1.43 & 3.92 \\
\textbf{ProtoKV} & 58.0 & 23.0 & 1.06 & 2.90\\
\bottomrule
\end{tabular}
\label{tab:rekv}
\end{table}

\section{Related Work}

\textbf{SVU with bounded memory.}
A common streaming baseline is sliding-window KV caching, which guarantees constant memory but can discard decisive cues when queries arrive after long and unpredictable delays. InfiniPot-V \cite{infinipotv} extends this regime with online token-retention/compression under a fixed cap, mitigating outright eviction yet still representing far history as a subset of token instances — a limitation our summary-state design directly targets. As a complementary approach, ReKV \cite{rekv} attempts to mitigate information loss by selectively recalling critical KV pairs, though its efficiency is often bounded by the requirement of a relatively large memory budget to sustain accuracy. Alternatively, ProVideLLM \cite{providellm} utilizes a multimodal interleaved cache to represent distant history through verbalized text tokens, though it necessitates specialized training for visual-text alignment. Parallel to cache-centric methods, VideoLLM-MoD \cite{videollm_mod} optimizes computational efficiency by dynamically skipping layer-wise processing for redundant visual tokens via a mixture-of-depths approach. 

\textbf{Offline Video Understanding and Long-Context VLMs.}
Long-context VLMs typically compress video tokens in offline settings where the full sequence is accessible. LongVU \cite{longvu} employs a training-based adaptive compression module, while DyCoke \cite{dycoke} offers a training-free alternative via importance-based pruning. However, these offline-centric designs are not optimized for the continuous update pressure of streaming SVU, where fixed memory budgets must sustain sparse, decisive evidence over indefinite horizons without full-sequence re-evaluation.

\textbf{KV Cache Compression in Large Language Models.}
KV cache management in LLMs primarily relies on sliding windows \cite{streamingllm} or importance-based eviction \cite{snapkv, h20}. While effective for discrete linguistic tokens, these token-selection strategies are often too brittle for video streams. As demonstrated in our experiments, such selection-only approaches fail to preserve sparse visual cues when subjected to long evidence delays and sustained post-evidence updates, a gap our summary-state approach directly addresses.

\section{Conclusion}
In this work, we address streaming video understanding under strict memory budgets, where decisive evidence can appear briefly but must remain usable after many subsequent updates before a query arrives. We propose ProtoKV, a constant-footprint KV memory that keeps an exact near-window cache while representing far history as an object-centric prototype bank with lightweight residual statistics. At query time, ProtoKV exposes this summary state through a bounded pseudo-token interface compatible with standard attention, keeping query-time computation controlled. Experiments on SVU benchmarks show that ProtoKV achieves competitive or improved accuracy under matched budgets and remains significantly more robust as query delay grows, with larger gains in the long-delay regime. These results highlight the value of summary-state far memory for stable streaming systems under sustained update pressure, and motivate future work on richer online summaries and broader SVU settings.
\section*{Acknowledgement}
We appreciate anonymous reviewers for providing constructive feedback and suggestions.
This work was supported by Samsung Electronics, 
Institute for Information \& communications Technology Planning \& Evaluation (IITP) grant funded by the Korea government (MSIT)(RS-2019-II190075,Artificial Intelligence Graduate School Support Program (KAIST)), and
National Research Foundation of Korea (NRF) grant funded by the Korea government (MSIT) (No.RS-2024-00340099).

\section*{Impact Statement}
This paper presents work whose goal is to advance the field of machine learning, specifically improving the memory and latency efficiency of streaming video understanding. By lowering deployment cost, this can broaden access to video-capable models in resource-constrained settings. There are many potential societal consequences of our work, none of which we feel must be specifically highlighted here.

\bibliography{citation}

@inproceedings{rvs,
  title={Flash-{VStream}: Efficient Real-Time Understanding for Long Video Streams},
  author={Zhang, Haoji and Wang, Yiqin and Tang, Yansong and Liu, Yong and Feng, Jiashi and Jin, Xiaojie},
  booktitle={Proceedings of the IEEE/CVF International Conference on Computer Vision},
  year={2025}
}

@article{providellm,
  title={Memory-efficient Streaming VideoLLMs for Real-time Procedural Video Understanding},
  author={Chatterjee, Dibyadip and Remelli, Edoardo and Song, Yale and Tekin, Bugra and Mittal, Abhay and Bhatnagar, Bharat and Camg{\"o}z, Necati Cihan and Hampali, Shreyas and Sauser, Eric and Ma, Shugao and others},
  journal={arXiv preprint arXiv:2504.13915},
  year={2025}
}

@inproceedings{videollm-online,
  title={Videollm-online: Online video large language model for streaming video},
  author={Chen, Joya and Lv, Zhaoyang and Wu, Shiwei and Lin, Kevin Qinghong and Song, Chenan and Gao, Difei and Liu, Jia-Wei and Gao, Ziteng and Mao, Dongxing and Shou, Mike Zheng},
  booktitle={Proceedings of the IEEE/CVF Conference on Computer Vision and Pattern Recognition},
  pages={18407--18418},
  year={2024}
}

@inproceedings{reST,
  title={Relational space-time query in long-form videos},
  author={Yang, Xitong and Chu, Fu-Jen and Feiszli, Matt and Goyal, Raghav and Torresani, Lorenzo and Tran, Du},
  booktitle={Proceedings of the IEEE/CVF Conference on Computer Vision and Pattern Recognition},
  pages={6398--6408},
  year={2023}
}

@article{videollm_mod,
  title={Videollm-mod: Efficient video-language streaming with mixture-of-depths vision computation},
  author={Wu, Shiwei and Chen, Joya and Lin, Kevin Qinghong and Wang, Qimeng and Gao, Yan and Xu, Qianli and Xu, Tong and Hu, Yao and Chen, Enhong and Shou, Mike Zheng},
  journal={Advances in Neural Information Processing Systems},
  year={2024}
}

@inproceedings{vllm,
  title={Efficient memory management for large language model serving with pagedattention},
  author={Kwon, Woosuk and Li, Zhuohan and Zhuang, Siyuan and Sheng, Ying and Zheng, Lianmin and Yu, Cody Hao and Gonzalez, Joseph and Zhang, Hao and Stoica, Ion},
  booktitle={Proceedings of the 29th symposium on operating systems principles},
  pages={611--626},
  year={2023}
}

@article{pq,
  title={Product quantization for nearest neighbor search},
  author={Jegou, Herve and Douze, Matthijs and Schmid, Cordelia},
  journal={IEEE transactions on pattern analysis and machine intelligence},
  volume={33},
  number={1},
  pages={117--128},
  year={2010},
  publisher={IEEE}
}

@inproceedings{shadowkv,
  title     = {{ShadowKV}: {KV} Cache in Shadows for High-Throughput Long-Context {LLM} Inference},
  author    = {Sun, Hanshi and Chang, Li-Wen and Bao, Wenlei and Zheng, Size and Zheng, Ningxin and Liu, Xin and Dong, Harry and Chi, Yuejie and Chen, Beidi},
  booktitle = {Proceedings of the 42nd International Conference on Machine Learning},
  pages     = {57355--57373},
  series    = {Proceedings of Machine Learning Research},
  volume    = {267},
  publisher = {PMLR},
  year      = {2025},
}

@article{llava-ov,
  title   = {{LLaVA}-{OneVision}: Easy Visual Task Transfer},
  author  = {Li, Bo and Zhang, Yuanhan and Guo, Dong and Zhang, Renrui and Li, Feng and Zhang, Hao and Zhang, Kaichen and Zhang, Peiyuan and Li, Yanwei and Liu, Ziwei and Li, Chunyuan},
  journal = {Transactions on Machine Learning Research},
  year    = {2025},
}

@article{qwen25,
  title={{Qwen2.5-VL} Technical Report},
  author={Bai, Shuai and Chen, Keqin and Liu, Xuejing and Wang, Jialin and Ge, Wenbin and Song, Sibo and Dang, Kai and Wang, Peng and Wang, Shijie and Tang, Jun and others},
  journal={arXiv preprint arXiv:2502.13923},
  year={2025}
}

@article{qwen2_,
  title={Qwen2-{VL}: Enhancing Vision-Language Model's Perception of the World at Any Resolution},
  author={Wang, Peng and Bai, Shuai and Tan, Sinan and Wang, Shijie and Fan, Zhihao and Bai, Jinze and Chen, Keqin and Liu, Xuejing and Wang, Jialin and Ge, Wenbin and Fan, Yang and Dang, Kai and Du, Mengfei and Ren, Xuancheng and Men, Rui and Liu, Dayiheng and Zhou, Chang and Zhou, Jingren and Lin, Junyang},
  journal={arXiv preprint arXiv:2409.12191},
  year={2024}
}

@inproceedings{ovobench,
  title={OVO-Bench: How Far is Your Video-LLMs from Real-World Online Video Understanding?},
  author={Niu, Junbo and Li, Yifei and Miao, Ziyang and Ge, Chunjiang and Zhou, Yuanhang and He, Qihao and Dong, Xiaoyi and Duan, Haodong and Ding, Shuangrui and Qian, Rui and others},
  booktitle={Proceedings of the IEEE/CVF Conference on Computer Vision and Pattern Recognition},
  pages={18902--18913},
  year={2025}
}

@article{streamingbench,
  title={Streamingbench: Assessing the gap for mllms to achieve streaming video understanding},
  author={Lin, Junming and Fang, Zheng and Chen, Chi and Wan, Zihao and Luo, Fuwen and Li, Peng and Liu, Yang and Sun, Maosong},
  journal={arXiv preprint arXiv:2411.03628},
  year={2024}
}

@article{infinipotv,
  title   = {{InfiniPot}-{V}: Memory-Constrained {KV} Cache Compression for Streaming Video Understanding},
  author  = {Kim, Minsoo and Shim, Kyuhong and Choi, Jungwook and Chang, Simyung},
  journal = {Advances in Neural Information Processing Systems},
  year    = {2025},
}

@inproceedings{vit,
  title     = {An Image is Worth 16x16 Words: Transformers for Image Recognition at Scale},
  author    = {Dosovitskiy, Alexey and Beyer, Lucas and Kolesnikov, Alexander and Weissenborn, Dirk and Zhai, Xiaohua and Unterthiner, Thomas and Dehghani, Mostafa and Minderer, Matthias and Heigold, Georg and Gelly, Sylvain and Uszkoreit, Jakob and Houlsby, Neil},
  booktitle = {International Conference on Learning Representations},
  year      = {2021},
}

@inproceedings{llama_vid,
  title={Llama-vid: An image is worth 2 tokens in large language models},
  author={Li, Yanwei and Wang, Chengyao and Jia, Jiaya},
  booktitle={European Conference on Computer Vision},
  pages={323--340},
  year={2024},
  organization={Springer}
}

@inproceedings{ego4d,
  title={Ego4d: Around the world in 3,000 hours of egocentric video},
  author={Grauman, Kristen and Westbury, Andrew and Byrne, Eugene and Chavis, Zachary and Furnari, Antonino and Girdhar, Rohit and Hamburger, Jackson and Jiang, Hao and Liu, Miao and Liu, Xingyu and others},
  booktitle={Proceedings of the IEEE/CVF conference on computer vision and pattern recognition},
  pages={18995--19012},
  year={2022}
}

@inproceedings{rekv,
  title={Streaming video question-answering with in-context video kv-cache retrieval},
  author={Di, Shangzhe and Yu, Zhelun and Zhang, Guanghao and Li, Haoyuan and Zhong, Tao and Cheng, Hao and Li, Bolin and He, Wanggui and Shu, Fangxun and Jiang, Hao},
  booktitle = {International Conference on Learning Representations},
  year={2025}
}

@inproceedings{streamingllm,
  title     = {Efficient Streaming Language Models with Attention Sinks},
  author    = {Xiao, Guangxuan and Tian, Yuandong and Chen, Beidi and Han, Song and Lewis, Mike},
  booktitle = {International Conference on Learning Representations},
  year      = {2024},
}

@article{h20,
  title={H2o: Heavy-hitter oracle for efficient generative inference of large language models},
  author={Zhang, Zhenyu and Sheng, Ying and Zhou, Tianyi and Chen, Tianlong and Zheng, Lianmin and Cai, Ruisi and Song, Zhao and Tian, Yuandong and R{\'e}, Christopher and Barrett, Clark and others},
  journal={Advances in Neural Information Processing Systems},
  volume={36},
  pages={34661--34710},
  year={2023}
}

@inproceedings{compressivetransformer,
  title     = {Compressive Transformers for Long-Range Sequence Modelling},
  author    = {Rae, Jack W. and Potapenko, Anna and Jayakumar, Siddhant M. and Lillicrap, Timothy P.},
  booktitle = {International Conference on Learning Representations},
  year      = {2020},
}

@article{snapkv,
  title={Snapkv: Llm knows what you are looking for before generation},
  author={Li, Yuhong and Huang, Yingbing and Yang, Bowen and Venkitesh, Bharat and Locatelli, Acyr and Ye, Hanchen and Cai, Tianle and Lewis, Patrick and Chen, Deming},
  journal={Advances in Neural Information Processing Systems},
  volume={37},
  pages={22947--22970},
  year={2024}
}

@article{longformer,
  title={Longformer: The long-document transformer},
  author={Beltagy, Iz and Peters, Matthew E and Cohan, Arman},
  journal={arXiv preprint arXiv:2004.05150},
  year={2020}
}

@inproceedings{prefixtuning,
  title     = {Prefix-Tuning: Optimizing Continuous Prompts for Generation},
  author    = {Li, Xiang Lisa and Liang, Percy},
  booktitle = {Proceedings of the 59th Annual Meeting of the Association for Computational Linguistics and the 11th International Joint Conference on Natural Language Processing (Volume 1: Long Papers)},
  pages     = {4582--4597},
  publisher = {Association for Computational Linguistics},
  doi       = {10.18653/v1/2021.acl-long.353},
  year      = {2021},
}

@inproceedings{longtail1,
  title     = {Long-tail Learning via Logit Adjustment},
  author    = {Menon, Aditya Krishna and Jayasumana, Sadeep and Rawat, Ankit Singh and Jain, Himanshu and Veit, Andreas and Kumar, Sanjiv},
  booktitle = {International Conference on Learning Representations},
  year      = {2021},
}

@article{longtail2,
  title={Balanced meta-softmax for long-tailed visual recognition},
  author={Ren, Jiawei and Yu, Cunjun and Ma, Xiao and Zhao, Haiyu and Yi, Shuai and others},
  journal={Advances in neural information processing systems},
  volume={33},
  pages={4175--4186},
  year={2020}
}

@article{transformer,
  title={Attention is all you need},
  author={Vaswani, Ashish and Shazeer, Noam and Parmar, Niki and Uszkoreit, Jakob and Jones, Llion and Gomez, Aidan N and Kaiser, {\L}ukasz and Polosukhin, Illia},
  journal={Advances in neural information processing systems},
  volume={30},
  year={2017}
}

@inproceedings{deepsort,
  title={Simple online and realtime tracking with a deep association metric},
  author={Wojke, Nicolai and Bewley, Alex and Paulus, Dietrich},
  booktitle={2017 IEEE international conference on image processing (ICIP)},
  pages={3645--3649},
  year={2017},
  organization={IEEE}
}

@inproceedings{dycoke,
  title={DyCoke: Dynamic Compression of Tokens for Fast Video Large Language Models},
  author={Tao, Keda and Qin, Can and You, Haoxuan and Sui, Yang and Wang, Huan},
  booktitle={Proceedings of the IEEE/CVF Conference on Computer Vision and Pattern Recognition},
  pages={18992--19001},
  year={2025}
}

@inproceedings{longvu,
  title     = {{LongVU}: Spatiotemporal Adaptive Compression for Long Video-Language Understanding},
  author    = {Shen, Xiaoqian and Xiong, Yunyang and Zhao, Changsheng and Wu, Lemeng and Chen, Jun and Zhu, Chenchen and Liu, Zechun and Xiao, Fanyi and Varadarajan, Balakrishnan and Bordes, Florian and Liu, Zhuang and Xu, Hu and Kim, Hyunwoo J. and Soran, Bilge and Krishnamoorthi, Raghuraman and Elhoseiny, Mohamed and Chandra, Vikas},
  booktitle = {Proceedings of the 42nd International Conference on Machine Learning},
  series    = {Proceedings of Machine Learning Research},
  volume    = {267},
  publisher = {PMLR},
  year      = {2025},
}

@inproceedings{videomme,
  title={Video-mme: The first-ever comprehensive evaluation benchmark of multi-modal llms in video analysis},
  author={Fu, Chaoyou and Dai, Yuhan and Luo, Yongdong and Li, Lei and Ren, Shuhuai and Zhang, Renrui and Wang, Zihan and Zhou, Chenyu and Shen, Yunhang and Zhang, Mengdan and others},
  booktitle={Proceedings of the IEEE/CVF Conference on Computer Vision and Pattern Recognition},
  pages={24108--24118},
  year={2025}
}

@inproceedings{mlvu,
  title={Mlvu: Benchmarking multi-task long video understanding},
  author={Zhou, Junjie and Shu, Yan and Zhao, Bo and Wu, Boya and Liang, Zhengyang and Xiao, Shitao and Qin, Minghao and Yang, Xi and Xiong, Yongping and Zhang, Bo and others},
  booktitle={Proceedings of the IEEE/CVF Conference on Computer Vision and Pattern Recognition},
  pages={13691--13701},
  year={2025}
}
\bibliographystyle{icml2026}

\newpage
\appendix
\onecolumn


\section{ProtoKV Method Details}
\label{sec:method_details}

\newcommand{\RETURN}{\STATE \textbf{return} }

\subsection{Streaming Memory State and Residual Representation}
\label{sec:method_state}

ProtoKV's bounded, query-agnostic streaming memory has a two-tier
state: a near ring buffer $B_{\mathrm{near}}$ storing the most recent
$W$ tokens' exact KV pairs, and a fixed-capacity far prototype bank
\[
  B_{\mathrm{far}} = \{\Pi_k\}_{k=1}^{K_{\max}},
  \quad
  \Pi_k = (c_k, p_k, n_k, R_k, \mu_k, \Sigma_k, \tau_k),
\]
where $(c_k, p_k)$ are key/value centers, $n_k$ is the prototype
mass, $R_k$ stores residual statistics, $(\mu_k, \Sigma_k)$ are
running spatial statistics, and $\tau_k$ records the stream position
of the most recently absorbed source token. Each prototype thus
summarizes many evicted tokens without retaining token instances.

\paragraph{Residual statistics.}
We implement $R_k$ as fixed-size Product Quantization (PQ) histograms
over key and value residuals:
\[
  R_k = \big(H_k^K, H_k^V, n_k^{\mathrm{res}}\big),
  \quad
  H_k^K, H_k^V \in \mathbb{N}^{G \times C},
\]
where $G$ is the number of PQ subquantizers and $C$ the codewords per
subquantizer. Entry $H_k^K[g, c]$ counts how often a key residual
assigned to prototype $k$ selects codeword $c$ in subspace $g$;
$H_k^V$ is defined analogously for value residuals. The scalar
$n_k^{\mathrm{res}}$ counts residual updates accumulated by prototype
$k$. Default values are $G = 8$ and $C = 16$
(\cref{sec:default_config}).

\paragraph{PQ codebook initialization.}
The PQ codebooks $\{C_g^K, C_g^V\}_{g=1}^G$ are initialized via
mini-batch $k$-means on a warm-up reservoir of residuals collected
during the first $T_{\mathrm{warm}}$ streaming steps, run separately
for each subspace and for key/value. Codebooks are shared across
prototypes within the same layer/head and kept fixed thereafter.


\subsection{Streaming Update Procedure}
\label{sec:method_update}

As the stream progresses, each visual token is appended to
$B_{\mathrm{near}}$; when the buffer overflows, the oldest token is
evicted and absorbed into $B_{\mathrm{far}}$ via continuity-aware
assignment (\cref{online_update} of the main paper), and the selected
prototype's centers, mass, recency, spatial statistics, and residual
histograms are updated. \cref{alg:streaming} gives the full
procedure.

\begin{algorithm}[!htb]
\caption{ProtoKV Streaming Update}
\label{alg:streaming}
\begin{algorithmic}[1]
\REQUIRE Near buffer $B_{\mathrm{near}}$, far prototype bank
  $B_{\mathrm{far}} = \{\Pi_k\}_{k=1}^{K_{\max}}$, incoming visual
  token $(K_i, V_i, s_i)$ at time $t$
\REQUIRE Near-window budget $W$, PQ codebooks
  $\{C_g^K, C_g^V\}_{g=1}^G$
\REQUIRE Assignment weights $\lambda_{\mathrm{sp}},
  \lambda_{\mathrm{idle}}$, idle threshold $T_{\mathrm{idle}}$,
  EMA rates $\alpha, \beta, \eta$
\STATE Append $(K_i, V_i, s_i, t)$ to $B_{\mathrm{near}}$
\IF{$|B_{\mathrm{near}}| > W$}
  \STATE Evict the oldest token $(K_e, V_e, s_e, t_e)$ from
    $B_{\mathrm{near}}$
  \IF{there exists an inactive prototype slot $k$}
    \STATE Initialize $\Pi_k$: $c_k \gets K_e$, $p_k \gets V_e$,
      $n_k \gets 1$, $\mu_k \gets s_e$, $\Sigma_k \gets I$,
      $\tau_k \gets t$, $R_k \gets (\mathbf{0}^{G \times C},
      \mathbf{0}^{G \times C}, 0)$
  \ELSE
    \STATE $k^\star \gets \arg\min_k \big[
      -\cos(K_e, c_k)
      + \lambda_{\mathrm{sp}} d_{\mathrm{Mah}}(s_e; \mu_k, \Sigma_k)
      + \lambda_{\mathrm{idle}} \mathbf{1}(t - \tau_k >
        T_{\mathrm{idle}}) \big]$
    \STATE $c_{k^\star} \gets (1-\alpha) c_{k^\star} + \alpha K_e$
    \STATE $p_{k^\star} \gets (1-\beta) p_{k^\star} + \beta V_e$
    \STATE $n_{k^\star} \gets n_{k^\star} + 1$,
      $\tau_{k^\star} \gets t$
    \STATE $\mu_{k^\star} \gets (1-\eta) \mu_{k^\star} + \eta s_e$
    \STATE $\Sigma_{k^\star} \gets (1-\eta) \Sigma_{k^\star} +
      \eta (s_e - \mu_{k^\star})(s_e - \mu_{k^\star})^\top$
    \STATE Update $R_{k^\star}$ from residuals
      $r^K = K_e - c_{k^\star}$, $r^V = V_e - p_{k^\star}$
  \ENDIF
  \STATE Apply \textsc{PrototypeMaintenance}($B_{\mathrm{far}}, t$)
\ENDIF
\STATE \RETURN $(B_{\mathrm{near}}, B_{\mathrm{far}})$
\end{algorithmic}
\end{algorithm}

After every absorption, ProtoKV applies three lightweight maintenance
operations (\cref{alg:maintenance}): \emph{aging} decays the mass of
idle prototypes; \emph{merging} combines redundant prototypes whose
key and value centers are sufficiently close; and \emph{recycling}
reinitializes prototype slots whose mass has decayed to zero,
reseeding them from a recent token in $B_{\mathrm{near}}$.

\begin{algorithm}[!htb]
\caption{ProtoKV Prototype Maintenance}
\label{alg:maintenance}
\begin{algorithmic}[1]
\REQUIRE Far prototype bank
  $B_{\mathrm{far}} = \{\Pi_k\}_{k=1}^{K_{\max}}$, current time $t$,
  near buffer $B_{\mathrm{near}}$
\REQUIRE Idle threshold $T_{\mathrm{idle}}$, decay factor $\gamma$,
  merge thresholds $\varepsilon_K, \varepsilon_V$
\FOR{each prototype $\Pi_k \in B_{\mathrm{far}}$}
  \IF{$t - \tau_k > T_{\mathrm{idle}}$}
    \STATE $n_k \gets \lfloor (1-\gamma) n_k \rfloor$
      \COMMENT{mass decay}
  \ENDIF
\ENDFOR
\FOR{each pair $(\Pi_i, \Pi_j)$ with $i \neq j$}
  \IF{$\| c_i - c_j \|_2 < \varepsilon_K$ \textbf{and}
      $\| p_i - p_j \|_2 < \varepsilon_V$}
    \STATE $c_i \gets \frac{n_i c_i + n_j c_j}{n_i + n_j}$,
      $p_i \gets \frac{n_i p_i + n_j p_j}{n_i + n_j}$
    \STATE $n_i \gets n_i + n_j$,
      $H_i^K \gets H_i^K + H_j^K$,
      $H_i^V \gets H_i^V + H_j^V$,
      $n_i^{\mathrm{res}} \gets n_i^{\mathrm{res}} + n_j^{\mathrm{res}}$
    \STATE $\mu_i \gets \frac{n_i \mu_i + n_j \mu_j}{n_i + n_j}$
    \STATE Mark $\Pi_j$ as inactive
      \COMMENT{merging}
  \ENDIF
\ENDFOR
\FOR{each prototype $\Pi_k \in B_{\mathrm{far}}$}
  \IF{$\Pi_k$ is empty (inactive or $n_k = 0$)}
    \STATE Select a recent token $(K_r, V_r, s_r)$ from
      $B_{\mathrm{near}}$
    \STATE Reinitialize $\Pi_k$: $c_k \gets K_r$, $p_k \gets V_r$,
      $n_k \gets 1$, $\mu_k \gets s_r$, $\Sigma_k \gets I$,
      $\tau_k \gets t$, $R_k \gets (\mathbf{0}^{G \times C},
      \mathbf{0}^{G \times C}, 0)$
      \COMMENT{recycling}
  \ENDIF
\ENDFOR
\STATE \RETURN $B_{\mathrm{far}}$
\end{algorithmic}
\end{algorithm}


\subsection{Query-Time Pseudo-Token Synthesis}
\label{sec:method_query_time}

At query time, ProtoKV exposes each prototype as $S$ pseudo-KV tokens
decoded from $R_k$, concatenates them with the near-window KV, and
feeds the result to standard attention. We treat each PQ histogram as
a factorized categorical distribution over code tuples. The smoothed
per-subspace probability for key residuals is
\begin{equation}
\label{eq:pq_prob}
  P^K_k(g, c) =
  \frac{H_k^K[g, c] + \epsilon}
       {\sum_{c'=1}^{C} H_k^K[g, c'] + C\epsilon},
\end{equation}
and a code tuple $z^K = (z^K_1, \ldots, z^K_G)$ scores
$\sum_{g} \log P^K_k(g, z^K_g)$. The top-$S$ tuples are obtained by
beam search over subquantizers (\cref{alg:decode}), avoiding the
$C^G$ exhaustive search. The decoded key residual for mode $s$ is
$\bar{r}^K_{k,s} = \mathrm{concat}_{g=1}^G C^K_{g, z^K_{k,s,g}}$;
value residuals are decoded analogously from $H_k^V$. By default we
decode key and value modes independently and pair them by rank,
yielding pseudo-tokens
\[
  \tilde{k}_{k,s} = c_k + \bar{r}^K_{k,s}, \qquad
  \tilde{v}_{k,s} = p_k + \bar{r}^V_{k,s}.
\]
This introduces a key--value independence approximation; a joint
histogram $H_k^{KV} \in \mathbb{N}^{G \times C \times C}$ would
decode paired modes directly at higher memory cost.

\paragraph{Bounded attention context.}
Concatenating pseudo-tokens from all $K_{\max}$ prototypes with the
near-window KV yields a query-time context of fixed length
$L_{\mathrm{context}} = W + K_{\max} \cdot S$, independent of stream
length. For each pseudo-token from prototype $k$, ProtoKV adds a
log-mass bias $b_k = \log n_k$ to the attention logit before softmax:
\begin{equation}
\label{eq:attn_logit}
  \ell_{k,s}
  = \frac{q^\top \tilde{k}_{k,s}}{\sqrt{d}} + \log n_k,
\end{equation}
shared across the $S$ pseudo-tokens of the same prototype. This
approximates the multiplicity effect of summarizing $n_k$ source
tokens, so a prototype's attention weight scales with accumulated
evidence. Near-window tokens carry zero bias.
\cref{alg:pseudo_decode} ties these steps together.

\begin{algorithm}[!htb]
\caption{DecodeTopSResidualModes}
\label{alg:decode}
\begin{algorithmic}[1]
\REQUIRE Residual histogram $H \in \mathbb{N}^{G \times C}$, PQ
  codebooks $\{C_g\}_{g=1}^G$, number of modes $S$, beam size $B$,
  smoothing $\epsilon$
\ENSURE Top-$S$ decoded residual modes $\{\bar{r}_s\}_{s=1}^S$
\STATE Initialize beam $\mathcal{B} \gets \{(0, \emptyset)\}$
  \COMMENT{log-score, partial code tuple}
\FOR{$g = 1$ to $G$}
  \STATE $P(g, c) \gets \dfrac{H[g, c] + \epsilon}
    {\sum_{c'=1}^{C} H[g, c'] + C\epsilon}$
  \STATE $\mathcal{B}' \gets \emptyset$
  \FOR{each $(a, z_{1:g-1}) \in \mathcal{B}$}
    \FOR{$c = 1$ to $C$}
      \STATE $\mathcal{B}' \gets \mathcal{B}' \cup
        \{ (a + \log P(g, c),\; [z_{1:g-1}, c]) \}$
    \ENDFOR
  \ENDFOR
  \STATE $\mathcal{B} \gets \mathrm{TopB}(\mathcal{B}')$
\ENDFOR
\STATE $\{z_s\}_{s=1}^S \gets \mathrm{TopS}(\mathcal{B})$
\FOR{$s = 1$ to $S$}
  \STATE $\bar{r}_s \gets \mathrm{concat}_{g=1}^G C_{g, z_{s,g}}$
\ENDFOR
\STATE \RETURN $\{\bar{r}_s\}_{s=1}^S$
\end{algorithmic}
\end{algorithm}

\begin{algorithm}[!htb]
\caption{ProtoKV Query-Time Pseudo-Token Decoding and Attention}
\label{alg:pseudo_decode}
\begin{algorithmic}[1]
\REQUIRE Query matrix $Q$, near buffer $B_{\mathrm{near}}$ with KV
  $(K_{\mathrm{near}}, V_{\mathrm{near}})$, far prototype bank
  $B_{\mathrm{far}} = \{\Pi_k\}_{k=1}^{K_{\max}}$
\REQUIRE PQ codebooks $\{C_g^K, C_g^V\}_{g=1}^G$, pseudo-tokens per
  prototype $S$, beam size $B$, smoothing $\epsilon$
\ENSURE Attention output $O$
\STATE $\tilde{K} \gets K_{\mathrm{near}}$,
  $\tilde{V} \gets V_{\mathrm{near}}$,
  $b \gets \mathbf{0}^{|K_{\mathrm{near}}|}$
\FOR{each active prototype
  $\Pi_k = (c_k, p_k, n_k, R_k, \ldots) \in B_{\mathrm{far}}$}
  \IF{$n_k^{\mathrm{res}} = 0$}
    \STATE $\bar{r}^K_{k,s}, \bar{r}^V_{k,s} \gets \mathbf{0}$ for
      all $s = 1, \ldots, S$
  \ELSE
    \STATE $\{\bar{r}^K_{k,s}\}_{s=1}^S \gets
      \textsc{DecodeTopSResidualModes}(H_k^K,
      \{C_g^K\}_{g=1}^G, S, B, \epsilon)$
    \STATE $\{\bar{r}^V_{k,s}\}_{s=1}^S \gets
      \textsc{DecodeTopSResidualModes}(H_k^V,
      \{C_g^V\}_{g=1}^G, S, B, \epsilon)$
  \ENDIF
  \FOR{$s = 1$ to $S$}
    \STATE $\tilde{k}_{k,s} \gets c_k + \bar{r}^K_{k,s}$,\quad
      $\tilde{v}_{k,s} \gets p_k + \bar{r}^V_{k,s}$
    \STATE $\tilde{K} \gets [\tilde{K};\; \tilde{k}_{k,s}]$,\quad
      $\tilde{V} \gets [\tilde{V};\; \tilde{v}_{k,s}]$,\quad
      $b \gets [b;\; \log n_k]$
  \ENDFOR
\ENDFOR
\STATE $A \gets Q \tilde{K}^\top / \sqrt{d} + b^\top$
\STATE $O \gets \mathrm{softmax}(A)\, \tilde{V}$
\STATE \RETURN $O$
\end{algorithmic}
\end{algorithm}

\section{Configuration and Hyperparameters}

\subsection{Default Hyperparameter Configuration}
\label{sec:default_config}

\cref{tab:default_config} lists the default hyperparameter values used in
all main experiments unless otherwise stated. The same defaults are
applied across all six benchmarks in this paper, without per-dataset
retuning. The only quantity varied across evaluation settings is the
external memory budget $|M|$, which determines $W$ and $K_{\max}$
according to the scaling rule described below.
 
\paragraph{Scaling rule for the memory budget.}
When the external memory budget $|M|$ changes, the near-window size $W$
and the prototype capacity $K_{\max}$ are scaled together to preserve
the default near--far ratio $W : (K_{\max} \cdot S) = 1 : 3$, with $S$
held fixed. All other hyperparameters in \cref{tab:default_config}
remain unchanged across budget settings.

\paragraph{Additional implementation constants.}
A few additional constants appear in our implementation but are not
critical to the main results: the warm-up window
$T_{\mathrm{warm}}$ used to collect residuals for PQ codebook
initialization (\cref{sec:method_details}) and the smoothing
constant $\epsilon$ used in the PQ probability estimate
(\cref{eq:pq_prob}). Empirical performance is not sensitive to small
perturbations of these constants within typical ranges; exact values are
specified in our open-source release.

\begin{table}[!htb]
\centering
\caption{Default hyperparameter configuration used in all main
experiments. The same values are used across all six benchmarks; only
$|M|$ varies across evaluation settings, which scales $W$ and $K_{\max}$
as described in the text.}
\label{tab:default_config}
\small
\begin{tabular}{llc}
\toprule
Group & Hyperparameter & Default \\
\midrule
\multirow{2}{*}{Memory state}
  & Near\,:\,Far split, $W : (K_{\max} \cdot S)$            & $1 : 3$ \\
  & Pseudo-tokens per prototype, $S$                        & $8$ \\
\midrule
\multirow{3}{*}{Assignment}
  & Spatial weight, $\lambda_{\mathrm{sp}}$                 & $0.1$ \\
  & Idle weight, $\lambda_{\mathrm{idle}}$                  & $0.01$ \\
  & Idle threshold, $T_{\mathrm{idle}}$                     & $120$ \\
\midrule
\multirow{2}{*}{Update (EMA)}
  & Key/value center rate, $\alpha = \beta$                 & $0.05$ \\
  & Spatial state rate, $\eta$                              & $0.05$ \\
\midrule
\multirow{2}{*}{Maintenance}
  & Idle decay rate, $\gamma$                             & $0.05$ \\
  & Key/value merge thresholds, $(\varepsilon_K, \varepsilon_V)$ & $(0.20, 0.25)$ \\
\midrule
\multirow{3}{*}{Residual statistics}
  & PQ subquantizers, $G$                                   & $8$ \\
  & Codewords per subquantizer, $C$                         & $16$ \\
  & Beam size, $B$                         & $4S$ \space  $(=32)$ \\
\bottomrule
\end{tabular}
\end{table}

\subsection{Hyperparameter Sensitivity}
\label{sec:hp_sensitivity}

ProtoKV exposes hyperparameters in three groups:
(i) \emph{budget allocation}, which divides the query-time token budget
$|M| = W + K_{\max} \cdot S$ between the near window and the far prototype
bank;
(ii) \emph{assignment and refresh} ($\lambda_{\mathrm{sp}}$,
$\lambda_{\mathrm{idle}}$, $T_{\mathrm{idle}}$), which control prototype
continuity and staleness behavior; and
(iii) \emph{update and maintenance}---the EMA rates $(\alpha, \beta)$ and
the idle-decay rate $\gamma$---which control long-run bank stability.
We fix a single default configuration (\cref{sec:default_config}) for all
main results and vary only the external token budget when reporting
scaling trends, so reported differences reflect the memory mechanism
rather than per-dataset retuning. This subsection evaluates ProtoKV's
sensitivity around this default operating point.

\paragraph{Budget allocation.}
Under a fixed total budget $|M| = 24\text{k}$, we vary the near--far split
by adjusting $W$ versus $K_{\max} \cdot S$ while holding
$|M| = W + K_{\max} \cdot S$ constant. We also vary the per-prototype
pseudo-token count $S$ with $K_{\max} \cdot S$ fixed. Results on
Qwen2.5-VL-7B and RVS-Ego are reported in
\cref{tab:budget_allocation}. Performance is similar across balanced
splits, with only the extreme near-heavy setting ($3{:}1$) degrading at
long delay because the far-memory budget becomes too small to retain
delayed evidence. The pseudo-token count $S$ is more influential, as it
directly controls the trade-off between the prototype count $K_{\max}$
and per-prototype representation granularity under the fixed budget; we
find $S=8$ to balance the two. Our default $1{:}3$ split was chosen to
match the InfiniPot-V comparison setting rather than tuned per benchmark.

\begin{table}[!htb]
\centering
\caption{Sensitivity to budget allocation under fixed $|M|=24\text{k}$
on Qwen2.5-VL-7B / RVS-Ego. Accuracy is reported at query delays
$\Delta=0$ and $\Delta=30$ minutes. Top: varying the near-window vs.\
far-memory split under a fixed total budget. Bottom: varying the
pseudo-token count $S$ per prototype with $K_{\max} \cdot S$ held fixed.}
\label{tab:budget_allocation}
\small
\begin{tabular}{lcc}
\toprule
$W : (K_{\max} \cdot S)$ & Acc.\ @ $\Delta=0$ & Acc.\ @ $\Delta=30$ \\
\midrule
$1{:}5$            & 61.2 & 48.5 \\
$1{:}3$ (default)  & 61.0 & 48.2 \\
$1{:}2$            & 60.6 & 48.0 \\
$3{:}1$            & 61.0 & 43.2 \\
\midrule
$S=4$              & 57.7 & 46.0 \\
$S=8$ (default)    & 61.0 & 48.2 \\
$S=16$             & 58.8 & 46.7 \\
\bottomrule
\end{tabular}
\end{table}

\paragraph{Update and maintenance coefficients.}
We additionally perturb each key update and maintenance coefficient by
$0.5\times$ and $2\times$ its default value on both RVS-Ego and RVS-Movie,
keeping all other settings at default. \cref{tab:coef_sweep} reports the
resulting accuracy changes relative to the default. Across all swept
coefficients, the changes remain modest, staying within $0.9$ points of
the default on both datasets. This indicates that ProtoKV is not highly
sensitive to these coefficients around its default setting, supporting
our use of the same defaults across all six benchmarks without
per-domain retuning.

\begin{table}[!htb]
\centering
\caption{Sensitivity to update and maintenance coefficients on RVS-Ego
and RVS-Movie under the default budget and Qwen2.5-VL-7B. Entries are
accuracy changes (points) relative to the default setting at
$\Delta = 0$. Default accuracy at $\Delta = 0$ is $61.0$ on RVS-Ego and
$53.3$ on RVS-Movie.}
\label{tab:coef_sweep}
\small
\begin{tabular}{lcccc}
\toprule
& \multicolumn{2}{c}{RVS-Ego} & \multicolumn{2}{c}{RVS-Movie} \\
\cmidrule(lr){2-3} \cmidrule(lr){4-5}
Param (default) & $\times 0.5$ & $\times 2$ & $\times 0.5$ & $\times 2$ \\
\midrule
$\alpha, \beta$ ($0.05, 0.05$)    & $-0.3$ & $-0.5$ & $-0.4$ & $-0.6$ \\
$\lambda_{\mathrm{sp}}$ ($0.1$)   & $-0.4$ & $-0.8$ & $-0.5$ & $-0.4$ \\
$\lambda_{\mathrm{idle}}$ ($0.01$) & $-0.2$ & $-0.3$ & $-0.3$ & $-0.8$ \\
$\gamma$ ($0.05$)                 & $-0.7$ & $-0.8$ & $-0.2$ & $-0.5$ \\
$T_{\mathrm{idle}}$ ($120$)       & $-0.2$ & $-0.9$ & $-0.4$ & $-0.4$ \\
\bottomrule
\end{tabular}
\end{table}

\section{Experimental Setting Details}

\subsection{Dataset Selection for Query-Delay Sweep}
\label{sec:dataset_selection}

\paragraph{Delayed-query protocol and validity constraint.}
Ideally, $t_0$ would be the last-seen timestamp of decisive evidence
for each query; the SVU benchmarks evaluated here do not annotate this
quantity, so we follow standard practice and use the dataset-provided
query timestamp as a proxy for $t_0$. For a video of length $L$, a
delayed query is valid only if $t_0 + \Delta \le L$. To ensure fair
comparisons across different $\Delta$ values, we evaluate on the
intersection subset of queries that remain valid for all $\Delta$ in
the sweep, so that every delay point is computed on the same set of
queries.
 
\paragraph{Retrospective query types per benchmark.}
Shifting the query time from $t_0$ to $t_q$ can alter the ground truth
for query types that depend on what is happening at the moment the
query is posed. We therefore restrict the sweep to retrospective query
types whose answers are determined by past video content and are
invariant to the shift, using each benchmark's native task taxonomy:
 
\begin{itemize}
  \setlength{\itemsep}{0pt}
  \item \textbf{RVS-Ego and RVS-Movie.} RVS \cite{rvs} categorizes queries into five answer types; we
    select the three types whose ground truth is determined by past
    events: \emph{whether something happened}, \emph{order judging},
    and \emph{what-event-order}.
  \item \textbf{OVO-Bench.} OVO-Bench \cite{ovobench} categorizes queries
    into three major types: forward active responding, real-time
    visual perception, and backward-tracing. We use only the
    \emph{backward-tracing} subset, whose answers refer to past
    evidence that remains constant regardless of when the query is
    posed.
  \item \textbf{StreamingBench.} StreamingBench \cite{streamingbench} is not included in the delay sweep.
    Its main benchmark setting in \cref{tab:main_benchmark} restricts to
    real-time visual understanding queries (per the benchmark's own
    protocol), which by construction depend on the moment the query is
    posed and are not invariant to a query-time shift.
\end{itemize}
 
\paragraph{Included vs.\ excluded examples.}
\cref{tab:delay_sweep_examples} gives representative included and excluded queries for the benchmarks used in the delay sweep experiments.
 
\begin{table}[!htb]
\centering
\caption{Examples of included and excluded queries for the
query-delay sweep, paired by benchmark to highlight the contrast.
Included queries are retained because their ground truth is determined
by past video content and is invariant to a query-time shift; excluded
queries reference the current moment and would change ground truth
under the shift.}
\label{tab:delay_sweep_examples}
\small
\begin{tabular}{p{0.13\linewidth} p{0.38\linewidth} p{0.38\linewidth}}
\toprule
Benchmark & Included (retrospective) & Excluded (current-moment) \\
\midrule
RVS-Ego &
``Did the person wash the vegetables before turning on the stove?''
\textit{(Order Judging)} &
``What is happening right now? Summarize the current scene.''
\textit{(Scene Summary)} \\
\addlinespace
OVO-Bench &
``Earlier in the video, which object did the person pick up?''
\textit{(Backward-tracing)} &
``What color is the object the person is holding now?''
\textit{(Real-time perception)} \\
\bottomrule
\end{tabular}
\end{table}

\subsection{Implementation Setup}

This subsection documents two parts of the experimental setup that
are needed to interpret the matched-budget comparisons:
(i) the per-baseline allocation of the memory budget $|M|$, and
(ii) the raw (pre-compression) visual context length under each
backbone's streaming pipeline.
 
\paragraph{Baseline configurations.}
All baselines are evaluated under the same benchmark protocol, video
preprocessing pipeline, backbone, and total memory budget $|M|$; the
memory mechanism is the only intended source of difference between
methods.
 
\begin{itemize}
  \setlength{\itemsep}{0pt}
  \item \textbf{SWA} (sliding-window attention) uses the entire memory
    budget as a single sliding window: given $|M|$, SWA retains the
    most recent $|M|$ tokens' exact KV pairs and truncates everything
    earlier. No additional hyperparameters are introduced.
  \item \textbf{InfiniPot-V} follows its original continual-compression
    setting. The total budget $|M|$ is split into a recent KV buffer
    of size $|M| - |C|$ and a far-memory cache of size $|C|$ with
    $|C|/|M| = 0.75$, where $|C|$ denotes the retained far-memory KV
    size. The update parameters $\alpha = 0.5$ and $\gamma = 0.125$
    are used as in the original work; we do not retune these
    parameters for any of our evaluation settings.
\end{itemize}
 
ProtoKV's own configuration is reported separately in
\cref{sec:default_config} and is not repeated here.
 
\paragraph{Raw sequence lengths.}
\label{sec:raw_seq_length}
For a delayed query at shifted time $t_q = t_0 + \Delta$, the
relevant pre-compression quantity is the raw visual prefix
accumulated up to $t_q$. We report this quantity for each backbone
so that the effective compression ratio against $|M|$ is explicit.
 
\textbf{LLaVA-OV setup.}
We sample the video stream at $0.5$\,fps and produce $196$ visual
tokens per sampled frame, yielding a token arrival rate of roughly
$98$ tokens per second of video. Under this rate, the maximum raw
visual prefix reached over the evaluated benchmarks is approximately
$352{,}800$ tokens on RVS-Ego, $198{,}200$ tokens on RVS-Movie, and
$205{,}600$ tokens on OVO-Bench.
 
\textbf{Qwen2.5-VL setup.}
We follow the same model-side sampling configuration as InfiniPot-V,
which caps the raw visual context at $49{,}920$ tokens before any
compression. The Qwen2-VL-7B configuration used in Table 3 and Table 10 follows the same setup.
 
In both setups the raw context is then compressed to the budget $|M|$
by each bounded-memory method, so the effective compression factor
scales with both stream length up to $t_q$ and the chosen $|M|$. The
LLaVA-OV setting represents the more aggressive compression regime in
our experiments, while the Qwen2.5-VL setting operates at a fixed raw
ceiling regardless of stream length.

\subsection{Online Update Overhead}
\label{sec:overhead_appendix}

The main paper reports a per-frame online update cost of $33.8$\,ms,
about $30.8\%$ of the total per-frame processing time, measured on
Qwen2.5-VL-7B with a single RTX 5090 GPU while streaming the RVS-Ego
benchmark. This subsection provides the measurement setup and a
per-stage breakdown that supports that number.
 
\paragraph{Setup.}
We profile ProtoKV during stream ingestion (i.e., outside of any
query). Each iteration of the streaming loop encodes one sampled frame
into visual tokens, appends them to the near-window KV buffer, and
applies ProtoKV's online update (eviction, continuity-aware
assignment, prototype centers/mass/recency update, and residual
histogram update) for any tokens that fall outside the near window.
Per-iteration latencies are averaged over a representative window of
frames after a warm-up phase, with CUDA synchronization between stages
to attribute time correctly. All measurements are taken under the
default configuration listed in \cref{sec:default_config}.
 
\paragraph{Where the update time goes.}
Within the $33.8$\,ms per-frame update, the two largest contributors
are the continuity-aware assignment and the prototype update
(centers, mass, recency, and residual-histogram updates), each
accounting for a substantial fraction of the total update time on this setup.
The remaining maintenance routine is comparatively inexpensive because
the prototype bank is small ($K_{\max}$ on the order of a few thousand
under the default budget) and most maintenance steps reduce to
constant-time checks. We do not separately profile the upstream
visual encoder and KV append stages here; they are unchanged from the
backbone's default streaming pipeline.
 
\paragraph{Why this does not affect query-time TTFT.}
The $33.8$\,ms update cost is incurred during stream ingestion rather
than at query time, and the streaming loop runs continuously
regardless of whether a query is pending. In sampled-frame SVU
pipelines (e.g., $0.5$\,fps for LLaVA-OV in our setup,
\cref{sec:raw_seq_length}), the inter-frame interval is on the order
of seconds, which is much larger than the per-frame update cost.
ProtoKV's update can therefore be overlapped with this inter-frame
interval, and the work that contributes to query-time TTFT is just
the bounded query-time attention over $W + K_{\max} \cdot S$ tokens,
not the update cost. This separation is what allows ProtoKV to
maintain low query-time TTFT even under continuous stream ingestion.
\section{Additional Experiments}

\subsection{RoPE Position Anchor Comparison}
\label{sec:rope_appendix}

The main paper states that each prototype $k$ shares a single rotary
position $\tau_k$ across its $S$ pseudo-tokens, defined as the original
stream position of the most recently absorbed source token assigned to
that prototype. This subsection compares this default choice against
two natural alternatives that aggregate the positions of the source
tokens that have been absorbed into prototype $k$ over time:
 
\begin{itemize}
  \setlength{\itemsep}{0pt}
  \item \textbf{First}: $\tau_k$ is set to the stream position of the
    \emph{first} source token absorbed into the prototype, i.e.,
    the position at which the prototype was initialized.
  \item \textbf{Average}: $\tau_k$ is set to the running mean of the
    stream positions of all source tokens absorbed into the prototype.
  \item \textbf{Most recent (default)}: $\tau_k$ is refreshed to the
    stream position of the most recently absorbed source token, as
    described in the main paper.
\end{itemize}
 
In all three variants, the $S$ pseudo-tokens of prototype $k$ share
the same $\tau_k$, and near-window tokens retain their original
positions; only the choice of which absorbed position $\tau_k$ tracks
changes.
 
\cref{tab:rope_anchor} reports accuracy under the three variants on
Qwen2.5-VL-7B / RVS-Ego under the default memory budget. The
recency-based anchor outperforms both alternatives, and the gap
widens as the anchor reaches further into the past. The first-token
anchor performs worst, consistent with the intuition that the
position at which a prototype was \emph{initialized} becomes
progressively stale as the prototype continues to absorb new tokens,
so by query time it no longer reflects when the supporting evidence
was most strongly present. The average anchor partially mitigates
this by drifting forward as new tokens are absorbed, but it still
under-weights the most recent supporting evidence. The most-recent
choice keeps $\tau_k$ close to the latest supporting evidence for
each prototype, which is the temporal context in which downstream
attention is most likely to use it.
 
\begin{table}[!htb]
\centering
\caption{Accuracy under three RoPE position-anchor choices for
$\tau_k$ on Qwen2.5-VL-7B / RVS-Ego under the default memory budget.
All three variants share the same per-prototype anchor across the $S$
pseudo-tokens; they differ only in which absorbed position the anchor
tracks.}
\label{tab:rope_anchor}
\small
\begin{tabular}{lc}
\toprule
$\tau_k$ choice & Accuracy (\%) \\
\midrule
First absorbed position    & 57.2 \\
Average absorbed position  & 59.6 \\
Most recent (default)      & \textbf{61.0} \\
\bottomrule
\end{tabular}
\end{table}

\subsection{MLVU Task-Type Breakdown}
\label{sec:task_breakdown}
 
\cref{tab:mlvu_breakdown} reports ProtoKV's per-task accuracy on the
seven MLVU task types under matched memory budget ($|M| = 6\text{k}$)
and Qwen2-VL-7B. The pattern is consistent with the design of
ProtoKV's summary-state far memory: prototypes preserve evidence about
localized state and visually distinct events well, but they absorb
repeated visually similar events into a single slot, which limits
absolute accuracy on counting-style tasks.
 
\begin{table}[!htb]
\centering
\caption{ProtoKV's per-task accuracy on MLVU under matched memory
budget ($|M| = 6\text{k}$) and Qwen2-VL-7B. Task abbreviations:
Order = Action Order, Count = Action Count, AR = Anomaly Recognition,
Needle = Needle QA, Ego = Ego Reasoning, TR = Topic Reasoning.
 ``Avg.'' is the official MLVU M-Avg score across the seven tasks.}
\label{tab:mlvu_breakdown}
\small
\begin{tabular}{lcccccccc}
\toprule
Task & Order & Count & AR & Needle & PlotQA & Ego & TR & Avg. \\
\midrule
Acc. (\%) & 54.8 & 36.9 & 68.5 & 79.7 & 74.4 & 65.9 & 86.3 & 68.9 \\
\bottomrule
\end{tabular}
\end{table}

\subsection{Dynamic Pseudo-Token Allocation}
In the main paper we fix the number of pseudo-tokens per prototype to a
constant $S$ across all prototypes, which gives every prototype the
same query-time representational capacity and keeps the analysis
uniform across the prototype bank. A natural question is whether
unequal per-prototype allocation under the \emph{same} total budget can
improve performance: some prototypes summarize relatively static
content with low residual variance, while others capture more dynamic
content with higher variance, and the latter may benefit from finer
exposure.
 
To probe this, we test a simple variance-aware allocation that keeps
the total pseudo-token count fixed at
$S_{\mathrm{tot}} = K_{\max} \cdot S$ (so the query-time context length
$L_{\mathrm{context}}$ is unchanged) and assigns each prototype $k$ a
share proportional to its residual variance:
\[
  S_k = S_{\mathrm{tot}} \cdot
  \frac{\mathrm{Var}(H_k)}{\sum_j \mathrm{Var}(H_j)}.
\]
On Qwen2.5-VL-7B and RVS-Ego under the default budget,
variance-aware allocation improves accuracy over the fixed-$S$
baseline (\cref{tab:dynamic_alloc}). The gain is larger at long delay,
consistent with the intuition that prototypes whose residual
distributions concentrate around a few dominant patterns benefit from
having more pseudo-tokens dedicated to those patterns when the query
arrives long after the supporting evidence.
 
\begin{table}[!htb]
\centering
\caption{Variance-aware pseudo-token allocation versus the fixed
default ($S = 8$ per prototype) under matched total budget
$S_{\mathrm{tot}} = K_{\max} \cdot S$ on Qwen2.5-VL-7B / RVS-Ego.}
\label{tab:dynamic_alloc}
\small
\begin{tabular}{lcc}
\toprule
Setting & Acc.\ @ $\Delta=0$ & Acc.\ @ $\Delta=30$ \\
\midrule
Fixed, $S = 8$ (default)        & 61.0 & 48.2 \\
Variance-aware, $S_k$ adaptive  & 61.7 & 50.0 \\
\bottomrule
\end{tabular}
\end{table}
 
The main paper retains the fixed-$S$ design because it instantiates
ProtoKV's summary-state interface with the smallest set of moving
parts: a uniform per-prototype representation, no extra allocation
policy, and a query-time context length that is determined purely by
the static configuration $(W, K_{\max}, S)$. The variance-aware
variant above is therefore best read as evidence that the
allocation budget itself is a useful additional degree of freedom for
summary-state SVU memory, rather than as a competing design that the
main results overlook. We leave a more thorough study of allocation
policies---including stability across streams, interaction with
prototype maintenance, and behavior under tighter overall budgets---to
future work.

\section{Qualitative Case Studies}
\label{sec:case_studies}

\newcommand{\casebox}[1]{%
  \par\smallskip\noindent
  \fbox{\begin{minipage}{0.96\linewidth}\small #1\end{minipage}}%
  \par\smallskip
}

\textbf{Case 1 (success, Needle QA): localized state retrieval.}

\casebox{
\textbf{Question.} What is the state of movement of the American toad
at the mouth of the den in the video? \\
\textbf{Ground truth.} Very little movement. \\
\textbf{ProtoKV.} Very little movement.
}
The relevant cue is spatially localized to a small region in the frame
and is consistently supported over a contiguous interval, so it can
remain represented in the far-memory summary without needing to be
disambiguated from many similar episodes. A small number of prototypes
carry enough mass and residual structure to preserve a localized state
across sustained post-evidence updates.
 
\textbf{Case 2 (success, Plot QA): narrative continuation across
distinct events.}

\casebox{
\textbf{Question.} At the beginning of the video, a woman in a red
coat and a man with a hat are talking in the car. What does the man
do after he opens the car door and leaves? \\
\textbf{Ground truth.} Make a phone call. \\
\textbf{ProtoKV.} Make a phone call.
}
The question requires linking an early scene (a conversation in the
car) to a later, visually distinct event (the man's action after
leaving the car). Because the early conversation and the later action
occur in different visual contexts, they tend to be absorbed into
different prototypes under continuity-aware assignment, and the
narrative thread connecting them can be reconstructed at query time
from the surviving prototype centers and their relative recency
$\tau_k$. ProtoKV identifies the correct continuation despite the
intervening updates between the two scenes.
 
\textbf{Case 3 (failure, Action Order): order errors among visually similar events.}

\casebox{
\textbf{Question.} Identify the option that corresponds to the order
of events as they occur in the video. \\
\textbf{Ground truth.} javelin throw $\rightarrow$ water sliding
$\rightarrow$ abseiling $\rightarrow$ making jewelry. \\
\textbf{ProtoKV.} water sliding $\rightarrow$ javelin throw
$\rightarrow$ abseiling $\rightarrow$ making jewelry.
}
ProtoKV preserves the positions of \emph{abseiling} and
\emph{making jewelry} but swaps the first two events. The two swapped
events are both fast-motion outdoor sports and are visually more
similar to each other than to the remaining events; under
continuity-aware assignment, such events are more likely to share a
prototype, in which case the prototype's recency anchor $\tau_k$
reflects only the most recently absorbed source token. The relative
order between events absorbed into different prototypes is preserved
through their $\tau_k$ values, but the relative order between events
absorbed into the same prototype is not. The same case also shows that
when the events \emph{are} visually distinguishable
(\emph{abseiling}, \emph{making jewelry}), ProtoKV preserves their
positions correctly, consistent with Case~2.
 
\textbf{Case 4 (failure, Action Count): repeated visually similar
events.}

\casebox{
\textbf{Question.} In this video, how many instances are there of the
``carving pumpkin'' action scene in total? \\
\textbf{Ground truth.} 5. \\
\textbf{ProtoKV.} 1.
}
The five carving instances are visually similar to each other, so
under continuity-aware assignment they are absorbed into the same
prototype rather than into distinct slots. The prototype's mass $n_k$
grows with each absorbed instance and reflects accumulated evidence,
but the boundaries between distinct episodes are not retained as
separate slots. As a result, the model has access to strong evidence
that the action occurred, but not to an explicit count of how many
distinct episodes it occurred in.


\end{document}